\documentclass[10pt,journal,compsoc]{IEEEtran}
\usepackage{graphicx}

\ifCLASSOPTIONcompsoc
  \usepackage[nocompress]{cite}
\else
  \usepackage{cite}
\fi
\usepackage{hyperref}       
\usepackage{url}            
\usepackage{epsfig}
\usepackage{graphicx}
\usepackage{booktabs}       
\usepackage{amsfonts}       
\usepackage{nicefrac}       
\usepackage{microtype}      
\usepackage{xcolor}         
\usepackage{amsmath}
\usepackage{amsfonts}
\usepackage{amssymb}
\usepackage{cleveref}
\usepackage{cancel}
\usepackage{amsthm}
\usepackage{wrapfig}
\usepackage{multirow}
\usepackage{subfig}
\usepackage{enumerate}
\newtheorem{prop}{Proposition}

\usepackage{enumitem}
\setlist[itemize]{leftmargin=*}
\ifCLASSINFOpdf
\else
\fi
\begin{document}
%
\title{On the Eigenvalues of Global Covariance Pooling for Fine-grained Visual Recognition}
%
%
%
%

\author{Yue~Song,~\IEEEmembership{Member,~IEEE,}
        Nicu~Sebe,~\IEEEmembership{Senior Member,~IEEE,}
        Wei~Wang,~\IEEEmembership{Member,~IEEE}
\IEEEcompsocitemizethanks{\IEEEcompsocthanksitem Yue Song, Nicu Sebe, and Wei Wang are with the Department
of Information Engineering and Computer Science, University of Trento, Trento 38123,
Italy.\protect\\
E-mail: \{yue.song, nicu.sebe, wei.wang\}@unitn.it}
\thanks{Manuscript received April 19, 2005; revised August 26, 2015.}}

%
%

\markboth{IEEE TRANSACTIONS ON PATTERN ANALYSIS AND MACHINE INTELLIGENCE}
{Shell \MakeLowercase{\textit{et al.}}: Bare Demo of IEEEtran.cls for Computer Society Journals}
%



\IEEEtitleabstractindextext{%
\begin{abstract}
The Fine-Grained Visual Categorization (FGVC) is challenging because the subtle inter-class variations are difficult to be captured. One notable research line uses the Global Covariance Pooling (GCP) layer to learn powerful representations with second-order statistics, which can effectively model inter-class differences. In our previous conference paper, we show that truncating small eigenvalues of the GCP covariance can attain smoother gradient and improve the performance on large-scale benchmarks. However, on fine-grained datasets, truncating the small eigenvalues would make the model fail to converge. This observation contradicts the common assumption that the small eigenvalues merely correspond to the noisy and unimportant information. Consequently, ignoring them should have little influence on the performance. To diagnose this peculiar behavior, we propose two attribution methods whose visualizations demonstrate that the seemingly unimportant small eigenvalues are crucial as they are in charge of extracting the discriminative class-specific features. Inspired by this observation, we propose a network branch dedicated to magnifying the importance of small eigenvalues. Without introducing any additional parameters, this branch simply amplifies the small eigenvalues and achieves state-of-the-art performances of GCP methods on three fine-grained benchmarks. Furthermore, the performance is also competitive against other FGVC approaches on larger datasets. Code is available at \href{https://github.com/KingJamesSong/DifferentiableSVD}{https://github.com/KingJamesSong/DifferentiableSVD}.

\end{abstract}

}

\maketitle

\IEEEdisplaynontitleabstractindextext

%
\IEEEpeerreviewmaketitle

\IEEEraisesectionheading{\section{Introduction}\label{sec:introduction}}

\IEEEPARstart{T}{he} Fine-Grained Visual Categorization (FGVC) aims to classify the subordinate categories from a given super-category (\emph{e.g.,} birds~\cite{WelinderEtal2010,van2015building} or dogs~\cite{khosla2011novel}). Compared with ordinary classification tasks, the FGVC is very challenging due to the subtle inter-class variations but significant intra-class differences. It has a multitude of real-world applications, such as image captioning~\cite{hendricks2016deep}, food recommendation~\cite{min2019food}, and image retrieval~\cite{pang2020solving}. Among the
approaches for the FGVC task, the Global Covariance Pooling (GCP) methods, which exploit the second-order feature statistics, have achieved impressive performances on common FGVC benchmarks and have attracted research interests from the computer vision community~\cite{li2017second,li2018towards,Yu_2018_ECCV,zheng2019learning,wang2020deep,song2021approximate}. The GCP method is a spectral meta-layer that computes the sample covariance from convolutional features and conducts the matrix normalization to obtain more powerful representations. In general, a standard GCP meta-layer first uses Singular Value Decomposition (SVD) to factorize the sample covariance into the eigenvalue and eigenvector matrices. Afterwards, the matrix logarithm~\cite{ionescu2015matrix} or matrix square root~\cite{lin2017improved} is performed on the eigenvalues for normalization.
Finally, the normalized covariance is used as the global representation and fed into the FC layer for exploitation of second-order statistics.


\begin{figure}[htbp]
    \centering
    \includegraphics[width=0.99\linewidth]{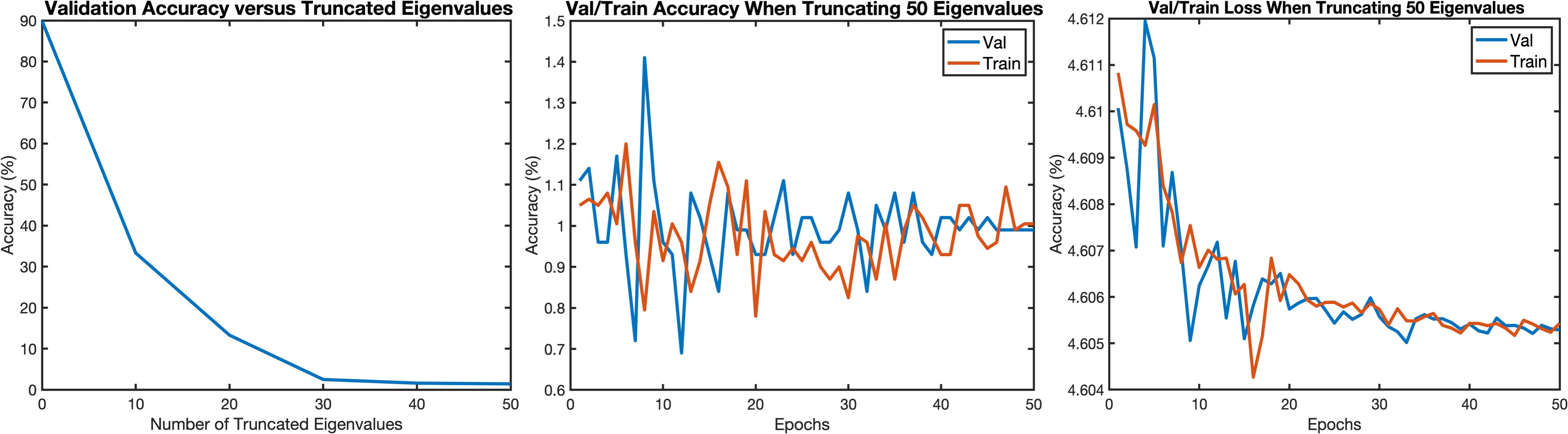}
    \caption{(\emph{Left}) Validation accuracy on Aircrafts versus the number of truncated eigenvalues. When the small eigenvalues are truncated, the performance drops drastically. After $30$ eigenvalues are truncated, the model cannot converge.
    (\emph{Middle and Right}) Validation/training accuracy and loss on Aircrafts versus training epochs when the last $50$ eigenvalues are truncated. The model fails to converge on the dataset.}
    \label{fig:acc_eig_cover}
\end{figure}

\begin{figure*}[t]
    \centering
    \includegraphics[width=0.99\linewidth]{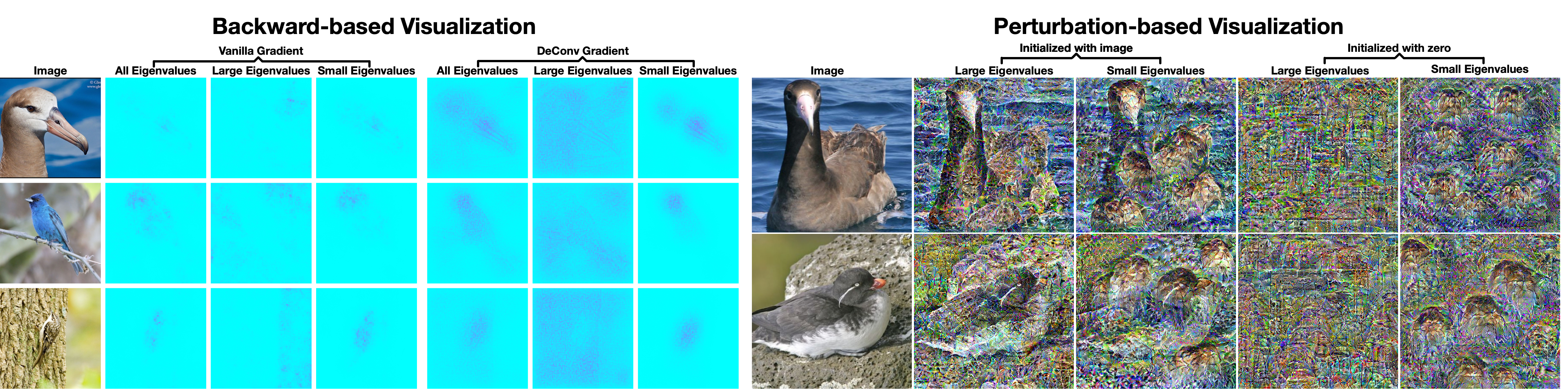}
    \captionsetup{font={small}}
    \caption{(\emph{Left}) Input responses of backward gradients to specific eigenvalues. Small eigenvalues (\emph{i.e.,} last 50 out of 256) highlight the salient class-discriminative regions, and they are coherent with the cases which use all eigenvalues. In contrast, the large eigenvalues (\emph{i.e.,} top 206) correspond to the background region. (\emph{Right}) Visualization of learned feature patterns that maximally activate the specific eigenvalues: small eigenvalues correspond to the class-specific features (\emph{e.g.,} feather, beak, and head), while the features associated with the large eigenvalues are not obviously class-relevant and human-interpretable. Zoom in for a better view.}
    \label{fig:back_perturb_visual}
\end{figure*}

In our previous conference paper~\cite{song2021approximate}, we conduct an investigation into the gradient smoothness of the SVD in the GCP layer. The SVD gradient involves the term $K_{ij}{=}\frac{1}{\lambda_{i}-\lambda_{j}}$ where $\lambda_{i}$ and $\lambda_{j}$ are eigenvalues. Since the covariance size of GCP is very large (\emph{i.e.,} 256${\times}$256), it is very likely to have many similar and small eigenvalues, \emph{i.e.,} $\lambda_{i}{\approx}\lambda_{j}$. This will cause the gradient term moves towards infinity (\emph{i.e.,} $K_{ij}{\rightarrow}\infty$) and trigger the numerical instability. To avoid this issue, a common practice is to truncate the small eigenvalues~\cite{lin2017improved,song2021approximate}, which could attain smoother gradients. We observe that when the small eigenvalues (\emph{i.e.,} last 50 out of 256) of the global covariance are truncated, the performance of GCP method on ImageNet~\cite{deng2009imagenet} could get improvements and the training will be more stable. However, as can be seen from Fig.~\ref{fig:acc_eig_cover}, on fine-grained recognition datasets truncating the small eigenvalues would make the model fail to converge. 
This intriguing phenomenon contradicts the common belief on the insignificant eigenvalues: \emph{as the eigenvalues decomposed by SVD are exponentially decayed, the majority of the matrix energy has been preserved well by the first few large eigenvalues. Truncating the small ones should not harm the compact representation of the data.} For a given matrix $\mathbf{P}$ and the truncated one $\mathbf{P}_{k}$ that keeps top-$k$ eigenvalues, according to Eckart-Young-Mirsky theorem~\cite{eckart1936approximation}, we have:
\begin{equation}
    ||\mathbf{P}-\mathbf{P}_{k}||_{\rm F}=\sqrt{\sigma_{k+1}^2+,\dots,+\sigma_{d}^{2}}
\end{equation}
where $\sigma_{i}$ is the $i$-th largest eigenvalue, and $d$ denotes the dimensionality. This is also known as the best low-rank approximation property of SVD. The theorem implies that truncating the insignificant eigenvalues still provides a very close approximation of the data (\emph{i.e.,} $||\mathbf{P}-\mathbf{P}_{k}||_{\rm F}{\approx}0$). In many practical applications such as image processing~\cite{sadek2012svd} and data mining~\cite{skillicorn2007understanding}, the noise of the data is usually hidden in the small eigenvalues. Therefore, truncating the small eigenvalues of large matrices has become a common practice.


An intuitive explanation is that the large eigenvalues capture statistics of principal directions along which the feature variances are large. Small eigenvalues, on the other hand, correspond to the features with smaller variance. However, for fine-grained recognition datasets, the inter-class feature differences (\emph{e.g.}, color of the bird head) are often very subtle. These features containing the classification clues are more likely to be encoded by the small eigenvalues. 
Unfortunately, due to the highly non-linear structure of CNNs, this assumption cannot be directly validated. 

We propose two visual explainability methods to diagnose the behavior of eigenvalues and associated eigenvectors. One methodology is to back-propagate the gradients of eigenvalues to the input image, which will highlight the distinct regions crucial to specific eigenvalues. The other technique is to perturb the image such that the final representation only embeds the large or small eigenvalues. The obtained images are in a deep-dreamed style~\cite{45507} and characterize the learned feature patterns. Through both quantitative evaluation and visual observation, the two aforementioned methods demonstrate that the small eigenvalues can effectively encode the discriminative and class-specific features. On the contrary, the large eigenvalues usually correspond to the background regions and fail in activating discriminative class-specific features compared with the small ones. As shown in Fig.~\ref{fig:back_perturb_visual}, for the small eigenvalues, the backward-based input responses consistently have salient points falling on the object, and the class-relevant features emerge repeatedly in the perturbation-based visualizations.



Our explainability methods attribute the decision cues of fine-grained classification to the small eigenvalues. Existing GCP methods usually use matrix square root normalization to reduce the magnitude of large eigenvalues~\cite{li2017second,li2018towards,yu2020toward}. However, the importance of small eigenvalues is not fully enhanced and remains less exploited. These observations naturally raise the following question: \emph{Can we increase the significance of small eigenvalues to make existing GCP models focus more on the semantically meaningful features and therefore to help improving their performance on fine-grained visual recognition?}

To solve this problem, we propose Scaling Eigen Branch (SEB), a general plug-in component for existing GCP models. It computes the exponential inverse of the covariance matrix that amplifies the relative significance of small eigenvalues. Then the covariance exponential inverse is used to generate a dynamic scaling factor and compensate for eigenvalues of the matrix square root. The backward gradient of the scaling factor helps the model to generate better-conditioned covariance matrices. The importance of small eigenvalues is thus magnified. Without introducing any additional parameters, the SEB improves the performances of GCP methods by $1.6\%$ on three popular fine-grained benchmarks, \emph{i.e.,} Caltech University Birds (Birds)~\cite{WelinderEtal2010}, Stanford Cars (Cars)~\cite{KrauseStarkDengFei-Fei_3DRR2013}, and FGVC Aircrafts (Aircrafts)~\cite{maji2013fine}. Moreover, the GCP method equipped with our proposed SEB also has very competitive performances against recent transformer-based FGVC approaches on some larger benchmarks including Stanford Dogs (Dogs)~\cite{khosla2011novel} and iNaturalist (iNats)~\cite{van2018inaturalist}. 

We summarize our contributions as follows:
\begin{itemize}
    \item We propose two visual explainability methods to diagnose the behavior of GCP eigenvalues, both of which demonstrate that small eigenvalues encode more class-specific features and make larger contributions to the decision-making process of GCP networks.
    \item We propose Scaling Eigen Branch (SEB), an add-on network branch to amplify the importance of small eigenvalues. Without bringing any extra parameters, the proposed SEB achieves the state-of-the-art performance of GCP methods on three fine-grained benchmarks. On large fine-grained datasets, our method also has comparable performance against other FGVC approaches.
\end{itemize}

This paper and our previous conference paper~\cite{song2021approximate} are connected but different. In the experiments of~\cite{song2021approximate}, we identify the peculiar behavior of GCP models on fine-grained recognition datasets, \emph{i.e.,} the model cannot converge without the last few eigenvalues. In this paper, we perform additional substantial work to explain the behavior and propose a plug-in network branch to improve the performances of GCP methods on the FGVC task. 

The rest of the paper is organized as follows: Sec.~\ref{sec:2} describes the related work in fine-grained classification, global covariance pooling, and visual explainability methods. Sec.~\ref{sec:3} presents our proposed methods to attribute the eigenvalues and Sec.~\ref{sec:4} introduces our SEB module that amplifies the small eigenvalues. Sec.~\ref{sec:5} provides the experimental results and in-depth analysis. Finally, Sec.~\ref{sec:6} summarizes the conclusions.

\section{Related Work}\label{sec:2}

\subsection{Fine-grained Visual Categorization}

The task of FGVC aims at distinguishing the subordinate categories of a given object category, \emph{e.g.,} birds~\cite{WelinderEtal2010,van2015building}, cars~\cite{KrauseStarkDengFei-Fei_3DRR2013}, and dogs~\cite{khosla2011novel}. Different from common classification scenarios, the FGVC is more challenging because the subtle inter-class variation needs to be captured. Existing works can be roughly divided into two groups: localization-based methods~\cite{berg2013poof,xie2013hierarchical,branson2014bird,huang2016part,ge2019weakly,huang2021stochastic} and representation-based approaches~\cite{lin2015bilinear,gao2016compact,lin2017improved,Kong_2017_CVPR,li2017second,li2018towards,gou2018monet,Yu_2018_ECCV,zheng2019learning,yu2020toward,du2020fine,song2021approximate}. The former group exploits the part annotations to localize the semantically-relevant regions and assist the classification, whereas the latter category targets learning more powerful representations with either attentive information or high-order statistics to model the subtle inter-class details. Recently, several vision transformer-based works have been proposed to tackle the challenge of FGVC~\cite{wang2021feature,liu2021transformer,he2021transfg}. In~\cite{wang2021feature}, the authors design a feature fusion strategy to increase the representation power of the ordinary vision transformer~\cite{dosovitskiy2020image} for the FGVC task. Thus, this work belongs to the representation-based category. The works of~\cite{liu2021transformer,he2021transfg} both design dedicated mechanisms to select the discriminative visual tokens or images patches to assist the FGVC, which can be considered as localization-based methods. The GCP methods that explore the second-order statistics are a particular kind of approach belonging to the representation-based category, and we will give a more detailed illustration in the following paragraph.

\subsection{Global Covariance Pooling }
In deep neural networks, Global Covariance Pooling (GCP) aims to explore the second-order statistics of convolutional features. DeepO$^{2}$P~\cite{ionescu2015matrix} is the first end-to-end global covariance pooling network. It formulates the theory of matrix back-propagation and demonstrates its effectiveness in visual recognition and segmentation tasks. Another pioneering work is B-CNN~\cite{lin2015bilinear} which proposes to aggregate the outer product of global features and perform element-wise power normalization. However, there exist two caveats in the two methods. First, the dimensionality of the covariance feature is too high, which significantly increases the parameters of the fully-connected layer and introduces the risk of overfitting. Secondly, the matrix logarithm normalization over-stretches the small eigenvalues and may not be effective enough. Based on the two pioneering works, the follow-up researches mainly proceed in three directions:

\begin{itemize}
    \item i) increase the representation power of global covariance by exploiting the distribution manifold or considering feature interactions in the convolutional layers ~\cite{li2017second,wang2017g2denet,zheng2019learning}; 
    \item ii) reduce the dimensionality of the covariance feature~\cite{Cui_2017_CVPR,gao2016compact,Kong_2017_CVPR}; 
    \item iii) seek for more efficient or effective matrix normalization schemes~\cite{lin2017improved,li2018towards,yu2020toward,zheng2019learning}. 
\end{itemize}
The second direction is closely related to the robustness of eigendecomposition. As pointed in~\cite{wang2021robust}, large matrices are more likely to have gradient explosion problems, but this has been solved by~\cite{song2021approximate}. It is also revealed that preserving the full dimension of covariance usually brings better performances. The second direction is therefore out of the scope of this paper. In this work, we mainly focus on the first and the last directions. To be more specific, we propose a plug-in component to improve the effectiveness of the global covariance feature. Also, our proposed component, together with the matrix square root normalization, can be viewed as a special matrix normalization scheme.



\subsection{Visual Explainability Methods}

Although the CNN has achieved remarkable success in many computer vision tasks, it is often considered a black-box model and suffers from weak interpretability. There have been many attempts to improve CNN’s explainability by visualizing the learned feature patterns or by highlighting the activation neurons. Existing visualization approaches can be roughly categorized into two families: backpropagation-based methods~\cite{simonyan2013deep,zeiler2014visualizing,springenberg2014striving,selvaraju2017grad,rebuffi2020there} and perturbation-based techniques~\cite{ribeiro2016should,fong2017interpretable,dabkowski2017real,petsiuk2018rise,fong2019understanding,zolna2020classifier}. The former category designs dedicated back-propagation rules to selectively highlight the input activations, whereas the latter perturbs or occludes the input image to maximally activate or confuse the classification predictions. Our work is closely related to both categories. More specifically, we use both backpropagation-based and perturbation-based techniques to provide the hint of the impact of eigenvalues. To the best of our knowledge, no similar work has been done in interpreting the impact of eigenvalues of GCP networks. 

\begin{table*}

\captionsetup{font={small}, justification=raggedright}
\caption{The average correlation coefficient and MAE between the input activation of the specific eigenvalues and the responses of all the eigenvalues. For the correlation coefficient, a higher number indicates the larger similarity, while the lower MAE implies smaller differences.}
\label{tab:similarity}
\centering
\resizebox{0.9\linewidth}{!}{
\begin{tabular}{c|c|ccc|ccc}\toprule  
\multirow{2}*{Gradient} & \multirow{2}*{Eigenvalue} & \multicolumn{3}{c|}{Correlation Coefficient} & \multicolumn{3}{c}{Mean Absolute Error}\\ \cline{3-8}
& & Birds~\cite{WelinderEtal2010} & Aircrafts~\cite{maji2013fine} & Cars~\cite{KrauseStarkDengFei-Fei_3DRR2013} & Birds~\cite{WelinderEtal2010} & Aircrafts~\cite{maji2013fine} & Cars~\cite{KrauseStarkDengFei-Fei_3DRR2013}\\
\hline
\multirow{2}*{Vanilla~\cite{simonyan2013deep}} &Large &0.06 &0.23 & 0.11 &3.5e-2 &3.3e-2 &3.5e-2\\  
&Small &\textbf{0.81} &\textbf{0.80} & \textbf{0.81} &\textbf{1.1e-2} &\textbf{1.0e-2} &\textbf{9.5e-3}\\
\hline
\multirow{2}*{DeConv~\cite{zeiler2014visualizing}}&Large & 0.61 & 0.70 & 0.75 &8.6e-2 &6.9e-2 &6.0e-2\\  
&Small & \textbf{0.89} & \textbf{0.91} & \textbf{0.93}& \textbf{3.4e-2} &\textbf{3.1e-2} &\textbf{2.4e-2}\\
\bottomrule
\end{tabular}
}

\end{table*} 

\section{Attributing the Eigenvalues} \label{sec:3}

In this section, we first revisit the procedure of global covariance pooling. Then, we introduce the two proposed visualization techniques in detail.

\subsection{Global Covariance Pooling Recap} 

Consider the reshaped convolutional feature ${\mathbf{X}}{\in}{\mathbb{R}^{d\times N}}$, where $d$ denotes the feature dimensionality (\emph{i.e.,} the number of channels) and $N$ represents the number of features (\emph{i.e.,} the product of spatial dimensions of features), a GCP meta-layer first computes the sample covariance matrix as:
\begin{equation}
    \mathbf{P}=\mathbf{X}\Bar{\mathbf{I}}\mathbf{X}^{T},\ \Bar{\mathbf{I}}=\frac{1}{N}(\mathbf{I}-\frac{1}{N}\mathbf{1}\mathbf{1}^{T})
    \label{covariance}
\end{equation}
where $\Bar{\mathbf{I}}$ represents the centering matrix, $\mathbf{I}$ denotes the identity matrix, and $\mathbf{1}$ is a column vector whose values are all ones, respectively. The sample covariance matrices are always symmetric positive semi-definite. Such matrices do not have any negative eigenvalues. Then the eigendecomposition is performed via SVD or eigenvalue decomposition (EIG): 
\begin{equation}
    \mathbf{P}=\mathbf{U}\mathbf{\Lambda}\mathbf{U}^{T},\ \mathbf{\Lambda}=\{\lambda_{1},\dots,\lambda_{d}\}_{\rm diag}
    \label{SVD}
\end{equation}
where $\mathbf{U}$ is the orthogonal eigenvector matrix, and $\mathbf{\Lambda}$ is the diagonal matrix in which the eigenvalues are sorted in a non-increasing order \emph{i.e.}, $\lambda_i {\geq} \lambda_{i+1}$. Afterwards, the matrix square root is conducted for normalization:
\begin{equation}
    \mathbf{Q}\triangleq\mathbf{P}^{\frac{1}{2}}=\mathbf{U}\mathbf{F}(\mathbf{\Lambda}) \mathbf{U}^{T}, \mathbf{F}(\mathbf{\Lambda})=\mathbf{\Lambda}^{\frac{1}{2}}=\{\lambda_{1}^{\frac{1}{2}},\dots,\lambda_{d}^{\frac{1}{2}}\}_{\rm diag}
    \label{matrix_power}
\end{equation}
where the normalized covariance matrix $\mathbf{Q}$ will be fed to the fully-connected layer. Besides the matrix square root normalization, there exist some other normalization schemes such as matrix logarithm~\cite{ionescu2015matrix} and rank-1 update~\cite{yu2020toward}. As the matrix square root is proved to amount to robust covariance estimation under the regularized MLE framework~\cite{li2017second}, this normalization is often preferred over other techniques. Existing state-of-the-art GCP methods are MPN-COV~\cite{li2017second} and iSQRT-COV~\cite{li2018towards}. The MPN-COV~\cite{li2017second} applies the above formulations for the covariance matrix computation. However, the standard SVD used in the MPN-COV is known to suffer from the gradient explosion problem as it involves the computation of $\frac{1}{\lambda_i - \lambda_j}$ where $i {\neq} j$. When $\lambda_i$ and $\lambda_j$ are very close or equal, the gradient will go to infinity. To avoid using SVD, the iSQRT-COV~\cite{li2018towards} proposes using Newton-Schulz iteration to directly derive the approximate matrix square root. Recently, the gradient explosion problem of SVD has been solved by~\cite{wang2021robust,song2021approximate} and a faster scheme of computing matrix square root is proposed in~\cite{song2022fast}. Throughout the experiments in this paper, we mainly investigate MPN-COV~\cite{li2017second} which is a standard GCP method that conducts explicit eigendecomposition, and we use the techniques of~\cite{song2021approximate} to compute the gradients.


We take ResNet-50 models of MPN-COV~\cite{li2017second} trained on the fine-grained benchmarks to diagnose the \textbf{peculiar} behavior of the eigenvalues.

\subsection{Backpropagation-based Methodology}

\subsubsection{Selective Back-propagation of Eigenvalues}

The core idea of our backward method is to manipulate the eigenvalue matrix defined in~\cref{SVD} and project the gradient back to the image. To visualize input activations to different eigenvalues, we could either abandon the small ones or the large ones from the eigenvalue matrix $\mathbf{\Lambda}$:
\begin{equation}
\begin{gathered}
    {\mathbf{\Lambda}}_{L}={\rm diag}\{\lambda_{1},\dots,\lambda_{t},0,\dots,0 \},\\
    {\mathbf{\Lambda}}_{S}={\rm diag}\{0,\dots,0,\lambda_{t+1},\dots,\lambda_{d}\}.
\end{gathered}
\end{equation}
where $t$ denotes the number of kept eigenvalues (\emph{i.e.,} top $206$ out of $256$), ${\mathbf{\Lambda}}_{L}$ represents the matrix that only preserves the large eigenvalues, and ${\mathbf{\Lambda}}_{S}$ is the matrix consisting of the small eigenvalues. The covariance matrix is usually of the size $256{\times}256$ and has $256$ non-negative eigenvalues in total. According to our observation on fine-grained datasets, for most covariance matrices, the last $50$ eigenvalues take less than $0.1\%$ energy of the matrix (\emph{i.e.,} $\frac{\sum_{i=t+1}^{d}\lambda_{i}}{\sum_{i=1}^{d}\lambda_{i}}{<}0.001$). Therefore, we categorize the last $50$ eigenvalues as the small eigenvalue $\mathbf{\Lambda}_{S}$ and the rest as the large eigenvalue $\mathbf{\Lambda}_{L}$.

Substituting the eigenvalue matrix $\mathbf{\Lambda}$ with ${\mathbf{\Lambda}}_{L}$ or ${\mathbf{\Lambda}}_{S}$ and back-propagating the gradients to the input could localize the salient regions that are crucial to the specific eigenvalues. 
To project the gradient back to the input, we need to consider the back-propagation rules of the GCP meta-layer and the standard CNN layers. For the sake of conciseness, we omit the back-propagation algorithm of the GCP meta-layer, and the readers are kindly referred to~\cite{ionescu2015training,li2017second,song2021approximate} for a detailed review. As for the gradient propagation of CNN, the main difficulty lies in the ReLU non-linearity. The vanilla backward gradient of ReLU~\cite{simonyan2013deep} is calculated by:
\begin{equation}
    {\rm ReLU}^{*}(p)=\begin{cases}
    1, & p>0\\
    0,  & p\leq0
    \end{cases}.
\end{equation}
where $p$ is the pixel-wise gradient back-propagated to ReLU. For the positive gradients, ReLU passes a constant to the next node in the computational graph, which loses the magnitude information and may not generate appealing visualizations. To resolve this issue, DeConv~\cite{zeiler2014visualizing} proposes an imputed back-propagation scheme:
\begin{equation}
    {\rm ReLU}_{\rm DeConv}^{*}(p)=\begin{cases}
    p, & p>0\\
    0,  & p\leq0
    \end{cases}.
\end{equation}
where the gradient magnitude is kept and more dynamic gradients are allowed. Compared with the vanilla gradient, DeConv gradient usually yields larger activated regions and more salient points, which could improve the visual appeal. We use both back-propagation rules to generate the input activations and they yield similar results.

Fig.~\ref{fig:back_perturb_visual} left displays several examples of input responses to all the eigenvalues, the large eigenvalues, and the small eigenvalues respectively. Obviously, for both rules, the visualizations of small eigenvalues are very similar to those of all eigenvalues. The salient points in their visualizations consistently fall on the objects. By contrast, the large eigenvalues mainly have activated neurons in the background and unimportant regions. More visualizations are provided in Supplementary Material.

\subsubsection{Quantitative Evaluation}

\noindent \textbf{Correlation Coefficient.} To quantitatively evaluate the importance of eigenvalues, we compute the average correlation coefficients between the input responses to the specific eigenvalues and those to all eigenvalues. The correlation coefficient is defined as:
\begin{equation}
    {\rm Corr}(A, B) {=} \frac{\sum_{m}\sum_{n}(A_{m,n}-\Bar{A})(B_{m,n}-\Bar{B})}{\sqrt{\sum_{m}\sum_{n}(A_{m,n}-\Bar{A})^{2}}\sqrt{\sum_{m}\sum_{n}(B_{m,n}-\Bar{B})^{2}}}
\end{equation}
where $A$ and $B$ are images to be compared, $\Bar{A}$ and $\Bar{B}$ are the mean intensity of the images, and the correlation coefficient ${\rm Corr}(\cdot,\cdot)$, which is normalized into the range of $[0,1]$, is used to assess their similarity. By measuring the similarity between the visualizations, we could understand what eigenvalues contribute more to the decision-making process. As the large eigenvalues using DeConv back-propagation rule have responses nearly everywhere on the image, the activated regions highly overlap with those using all the eigenvalues. Their correlation will be inevitably large. 

\noindent \textbf{Mean Absolute Error.} Alternatively, we could compute their Mean Absolute Error (MAE) to measure the statistical distance: 
\begin{equation}
    {\rm MAE}(A, B)=\frac{1}{M\times N}\sum_{m}\sum_{n}|{A_{m,n}-B_{m,n}}|
\end{equation}
where $M$ and $N$ denote the image width and height, respectively. The joint use of these two metrics can provide a comprehensive assessment on the similarities between the visualizations. Table~\ref{tab:similarity} compares the similarity between the input activations of specific eigenvalues and those of all the eigenvalues. For both metrics, the activated regions of small eigenvalues have a substantially larger similarity with the activated regions of all the eigenvalues. This demonstrates that the small eigenvalues preserve more semantically meaningful information and have greater contributions to the decisions of the GCP networks.

\begin{table}

\captionsetup{font={small}, justification=raggedright}
\caption{Classification accuracy (\%) using only subsets of eigenvalues during the inference stage.}
\label{tab:acc_subset}
\centering
\resizebox{0.99\linewidth}{!}{
\begin{tabular}{c|ccc}\toprule  
\multirow{2}*{Eigenvalue} & \multicolumn{3}{c}{Classification Accuracy (\%)} \\ 
\cline{2-4}
& Birds~\cite{WelinderEtal2010} & Aircrafts~\cite{maji2013fine} & Cars~\cite{KrauseStarkDengFei-Fei_3DRR2013}\\
\hline
$\lambda_{1},\dots,\lambda_{206}$ (top $206$) & 65.3 & 70.1 & 72.4 \\  
$\lambda_{1}+\lambda_{207},\dots,\lambda_{256}$ (last $50$) &\textbf{81.3} &\textbf{82.1} & \textbf{83.4}\\
$\lambda_{1}+\lambda_{217},\dots,\lambda_{256}$ (last $40$) &79.2 &80.7 &81.8 \\
$\lambda_{1}+\lambda_{227},\dots,\lambda_{256}$ (last $30$) &77.1 &78.5 &79.9 \\
\hline
$\lambda_{1},\dots,\lambda_{256}$ (all $256$) &84.3 &89.9 &91.7 \\
\bottomrule
\end{tabular}
}
\end{table} 

\noindent \textbf{Validation Accuracy.} Besides the analysis in the lens of visualization similarity, we also evaluate the impact of eigenvalues on the classification accuracy. Table~\ref{tab:acc_subset} presents the validation accuracy using different subsets of the eigenvalues in the inference stage. Since the last eigenvalues are very small in magnitude, merely using the small ones might poorly reconstruct the covariance representation and cannot output reasonable class predictions. Therefore, we also add the first eigenvalue $\lambda_{1}$ to keep the dominant vector space when testing the performance of small eigenvalues. As can be seen, even using the last $30$ eigenvalues has a higher validation accuracy than using the large ones. When the number of used eigenvalues is increased from $30$ to $50$, we can observe a steady performance improvement. This demonstrates that the small eigenvalues have more predictive power and a larger contribution to the decision-making process.

\subsection{Perturbation-based Methodology}

Instead of seeking for input responses to eigenvalues, another visualization method is to freeze the classifier weights and perturb the input images such that only specific eigenvalues and the associated eigenvectors are maximally activated. After keeping updating the image values for a certain number of iterations, the perturbed image will highlight the learned feature patterns of specific eigenvalues. For an input image, we first decompose the covariance matrix into two sub-matrices that embed only the large or small eigenvalues:
\begin{equation}
\mathbf{P}_{L}=\mathbf{U}\mathbf{\Lambda}_{L}\mathbf{U}^{T}, \mathbf{P}_{S}=\mathbf{U}\mathbf{\Lambda}_{S}\mathbf{U}^{T}
\end{equation}

\subsubsection{Visualization of Large Eigenvalues}

To visualize the feature patterns that correspond to the large eigenvalues, we could iteratively use the following loss function to perturb the image:
\begin{equation}
\begin{gathered}
    l_{1}=||\mathbf{M}-\mathbf{P}_{L}||_{\rm F}^{2}
    \label{explain_large}
\end{gathered}
\end{equation}
where $\mathbf{M}$ denotes the covariance matrix generated by the perturbed image, $l_{1}$ is the loss to optimize the image, and $||\cdot||_{\rm F}$ represents the Frobenius norm. The loss can push the perturbed image to generate the covariance composed only by large eigenvalues. Notice that the loss $l_{1}$ can be regarded as the Mean Square Error (MSE) loss between $\mathbf{P}_{L}$ and $\mathbf{M}$. 




\subsubsection{Visualization of Small Eigenvalues}

Deriving the perturbation loss for the small eigenvalues is more complex than the large ones. We cannot adopt the $||\mathbf{M}-\mathbf{P}_{S}||_{\rm F}^{2}$ loss as in~\cref{explain_large}. 
Since the eigenvalues $\mathbf{\Lambda}_{S}$ are sparse and exponentially smaller than the large ones, the reconstructed matrix $\mathbf{P}_{S}$ contains little energy of the original matrix $\mathbf{P}$ (often ${<}0.1\%$). It is infeasible for a covariance matrix $\mathbf{M}$ to have the identical eigenvalues as $\mathbf{P}_{S}$. Instead of pushing them to have the same eigenvalues, we enforce the eigenvalues of $\mathbf{M}$ and $\mathbf{P}_{S}$ to be highly correlated.
To achieve this goal, we introduce the following loss function:
\begin{equation}
    \begin{gathered}
    l_{2}=-||\mathbf{M}-\mathbf{P}_{L}||_{\rm F}^{2} + ||\mathbf{M}-\mathbf{P}_{S}||_{\rm F}^{2}
    \end{gathered}
\end{equation}
where the first term pushes $\mathbf{M}$ far from $\mathbf{P}_{L}$, and the second term makes sure that $\mathbf{M}$ stays close with $\mathbf{P}_{S}$.  The composition of the two losses can guarantee that $\mathbf{M}$ maximally activates the eigenvectors of $\mathbf{P}_{S}$ and has a highly correlated representation. Formally, we have the proposition as follows:

\begin{prop}
 The loss $l_{2}$ attains the minimum when the eigenvectors of $\mathbf{M}$ are $\mathbf{U}$, and the eigenvalues of $\mathbf{M}$ maximally correlate with $\mathbf{\Lambda}_{S}$ but have zero correlation with $\mathbf{\Lambda}_{L}$.
\end{prop}

\begin{proof}
    Relying on Frobenius inner product, the loss $l_{2}$ is decomposed by:
    
    \begin{equation}
        \begin{aligned}
        l_{2} &{=}-||\mathbf{M}-\mathbf{P}_{L}||_{\rm F}^{2} + ||\mathbf{M}-\mathbf{P}_{S}||_{\rm F}^{2}\\
        &{=} -\langle\mathbf{M}-\mathbf{P}_{L},\mathbf{M}-\mathbf{P}_{L}\rangle + \langle\mathbf{M}-\mathbf{P}_{S},\mathbf{M}-\mathbf{P}_{S}\rangle\\
        &{=}  2\langle\mathbf{M},\mathbf{P}_{L}\rangle {-} ||\mathbf{P}_{L}||_{\rm F}^{2}  {-} 2\langle\mathbf{M},\mathbf{P}_{S}\rangle {+} ||\mathbf{P}_{S}||_{\rm F}^{2}
        \end{aligned}
        \label{l2}
    \end{equation}
    
    Since both $||\mathbf{P}_{S}||_{\rm F}^{2}$ and $||\mathbf{P}_{L}||_{\rm F}^{2}$ are independent of $\mathbf{M}$, we can use a constant $C$ to represent the sum of these two terms:
    \begin{equation}
        \begin{aligned}
        l_{2} &= 2\langle\mathbf{M},\mathbf{P}_{L}\rangle - 2\langle\mathbf{M},\mathbf{P}_{S}\rangle + C\\
        &= -2\langle\mathbf{M}, \mathbf{P}_{S}-\mathbf{P}_{L}\rangle + C
        \end{aligned}
        \label{l2_decompose}
    \end{equation}
    For the inner product of a given matrix pair $\mathbf{A}$ and $\mathbf{B}$, Von Neumann’s trace inequality~\cite{mirsky1975trace,grigorieff1991note} tells:
    \begin{equation}
        |\langle\mathbf{A},\mathbf{B}\rangle|\leq \sigma_{1}(\mathbf{A})\sigma_{1}(\mathbf{B})+\dots+\sigma_{n}(\mathbf{A})\sigma_{n}(\mathbf{B})
        \label{trace_ineq}
    \end{equation}
    where $\sigma_{i}(\cdot)$ denotes the $i$-th eigenvalue. Injecting~\cref{trace_ineq} into~\cref{l2_decompose}, the loss $l_{2}$ can be re-expressed as:
    \begin{equation}
        \begin{aligned}
        l_{2}\geq - 2\sum_{i=1}^{d}\sigma_{i}(\mathbf{M})\sigma_{i}(\mathbf{P}_{S}-\mathbf{P}_{L}) + C
         \end{aligned}
         \label{l2_inequlity}
    \end{equation}
    Recall that $\mathbf{P}_{L}$ is composed by the large eigenvalues $\mathbf{\Lambda}_{L}{=}\{\lambda_{1},\dots,\lambda_{t},0,\dots,0 \}_{\rm diag}$ and $\mathbf{P}_{S}$ only embeds the small eigenvalues $\mathbf{\Lambda}_{S}{=}\{0,\dots,0,\lambda_{t+1},\dots,\lambda_{d}\}_{\rm diag}$. Eq. \ref{l2_inequlity} can be further formulated as:
    \begin{equation}
        \begin{aligned}
        l_{2} &\geq 2\sum_{i=1}^{t}\sigma_{i}(\mathbf{M})\lambda_{i} - 2\sum_{i=t+1}^{d}\sigma_{i}(\mathbf{M})\lambda_{i} + C
        \end{aligned}
        \label{l2_leq}
    \end{equation}
    Since $\mathbf{M}$ is positive semi-definitive, there will not exist any negative eigenvalues $\sigma_{i}(\mathbf{M})$. Therefore, the minimum of the first term is zero. It happens when the eigenvalues of $\mathbf{M}$ have a zero correlation with $\mathbf{\Lambda}_{L}$, \emph{i.e.,} the large eigenvalues of $\mathbf{M}$ are zero. 
    
    The second term actually measures the correlation between the last $(d-t)$ eigenvalues of $\mathbf{M}$ and $\mathbf{P}$. When the r.h.s. of~\cref{l2_leq} attains the minimum, the correlation will reach the maximum and $\mathbf{M}$ will maximally activate the eigenvectors associated with the small eigenvalues. The equality is taken when $\mathbf{M}$ also has the eigenvector matrix $\mathbf{U}$.
\end{proof}

This proposition shows that after the image is perturbed for certain iterations, \emph{i.e.,} the loss attains the minimum, $\mathbf{M}$ will only activate the learned feature patterns of the eigenvectors associated with small eigenvalues. But the eigenvectors that correspond to the large eigenvalues will not be encoded by $\mathbf{M}$. 
Fig.~\ref{fig:back_perturb_visual} right visualizes the learned feature patterns that correspond to the specific eigenvalues. Although mainly mid-level and low-level features are activated in the visualizations, the semantically meaningful patterns can be observed. For the small eigenvalues, the visualizations exhibit distinct class-relevant features. Take the $2^{nd}$ row as an example, the bird head, which characterizes the bird species, emerges repeatedly in the visualizations. On the contrary, the large eigenvalues do not activate obviously class-relevant and human-interpretable features. Unfortunately, this visualization technique can only be assessed through visual observation but cannot be evaluated by any reasonable quantitative metrics. We provide more visualization results in Supplementary Material.

\begin{figure*}[t]
    \centering
    \includegraphics[width=0.99\linewidth]{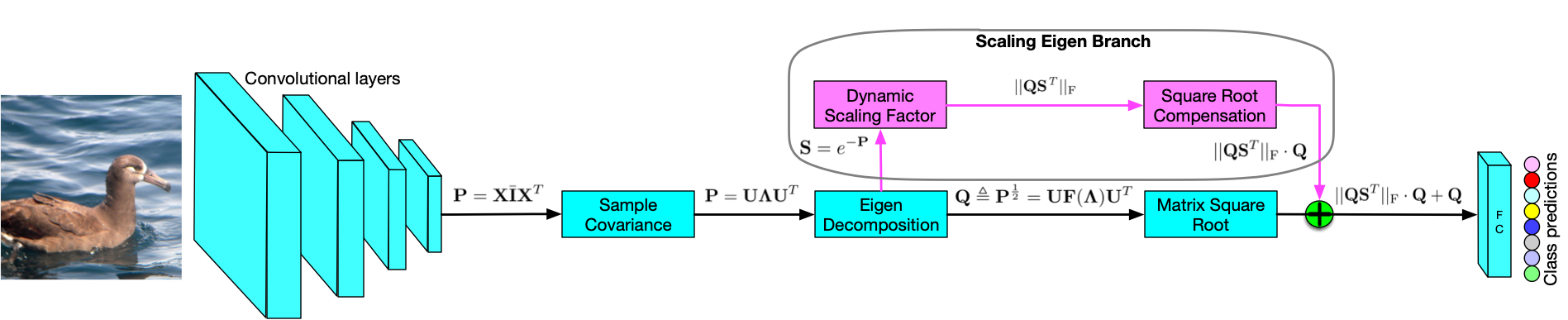}
    \captionsetup{font={small}}
    \caption{Overview of the standard global covariance pooling procedure (\emph{i.e.,} MPN-COV~\cite{li2017second} and iSQRT-COV~\cite{li2018towards}) and the proposed SEB component (indicated in pink). Our proposed SEB augments the representation power of matrix square root by magnifying the importance of small eigenvalues. Without introducing any additional parameters, the performances are significantly boosted.}
    \label{fig:SEB}
\end{figure*}

\section{Scaling Eigen Branch}
\label{sec:4}





The visualization results and quantitative analysis demonstrate that the small eigenvalues and associated eigenvectors are actually crucial to the network decisions as they capture rich semantic information and class-relevant feature patterns. This observation inspires us to propose SEB, a plug-in component dedicated to magnify the importance of small eigenvalues.

\subsection{Amplifying Small Eigenvalues by Scaling}

To magnify the importance of small eigenvalues, one straightforward approach is to increase their numerical values. However, the eigenvalues are sorted according to the varying extent of the feature along the eigenvector directions. Directly increasing the value of small eigenvalues could reverse the significance order and disturb the statistical information of covariance, which might make the model fail to converge. For the covariance matrix, it is important to maintain the significance order of the eigenvalues. Another intuitively possible method is to increase the relative importance of small eigenvalues (\emph{i.e.,} increase $\frac{\lambda_{d}}{\lambda_{1}}$) but maintain the significance order (\emph{i.e.} $\frac{\lambda_{i+1}}{\lambda_{i}}{\leq}1$ is always satisfied). This could be easily done by computing the matrix $p$-th root $\mathbf{P}^{\frac{1}{p}}$ instead of the square root $\mathbf{P}^{\frac{1}{2}}$. A small $\frac{1}{p}$ would punish the magnitude of large eigenvalues and magnify the significance of small ones. However, as pointed out in~\cite{li2017second}, only the matrix square root can amount to robust covariance estimation under the regularized MLE framework and approximately exploits Riemannian geometry under the Power-Euclidean metric~\cite{dryden2009non}. Besides the theoretical analysis, the matrix square root also achieves the best experimental performances among different power normalization schemes (\emph{e.g.}, $\mathbf{P}^{0.1}$ and $\mathbf{P}^{0.3}$)~\cite{li2017second}. All of these suggest that computing the covariance square root is indispensable for robust covariance pooling.

So far, the only practical approach seems to be properly transforming the covariance matrix such that the eigenvalues are amplified but the form of matrix square root still manifests. Since our covariance matrix is symmetric positive semi-definite (SPSD), it is appropriate to consider the geometry of SPSD manifold. In the space of SPSD manifold, the distance between covariance matrices is not measured by the traditional Euclidean metrics (\emph{i.e.,} $d(X,Y){=}||X{-}Y||^{2}_{\rm F}$). Instead, a non-Euclidean metric should be adopted. One commonly used metric is the Log-Euclidean metric~\cite{arsigny2007geometric} defined as:
\begin{equation}
    d_{L}^{2}(X,Y)=|| log(X)-log(Y)||_{\rm F}^{2}
\end{equation}

We can easily find that this metric is scale-invariant, \emph{i.e.,} scaling two covariance matrices keeps their distance unchanged. Therefore, it is appropriate to consider scaling the covariance square root by a factor ($>$1) such that the eigenvalues are amplified and the distance between two scaled covariance matrices does not change in the SPSD manifold. Notice that the scale-invariance property generally applies for the non-Euclidean metrics used in the SPSD manifold (\emph{e.g.,} affine-invariant Riemannian metric~\cite{pennec2006riemannian} and Stein divergence~\cite{sra2012new}). In the following paragraphs, we will illustrate how the covariance matrices get appropriately scaled in our proposed SEB (see also Fig.~\ref{fig:SEB}).

\begin{figure}
    \centering
    \includegraphics[width=0.8\linewidth]{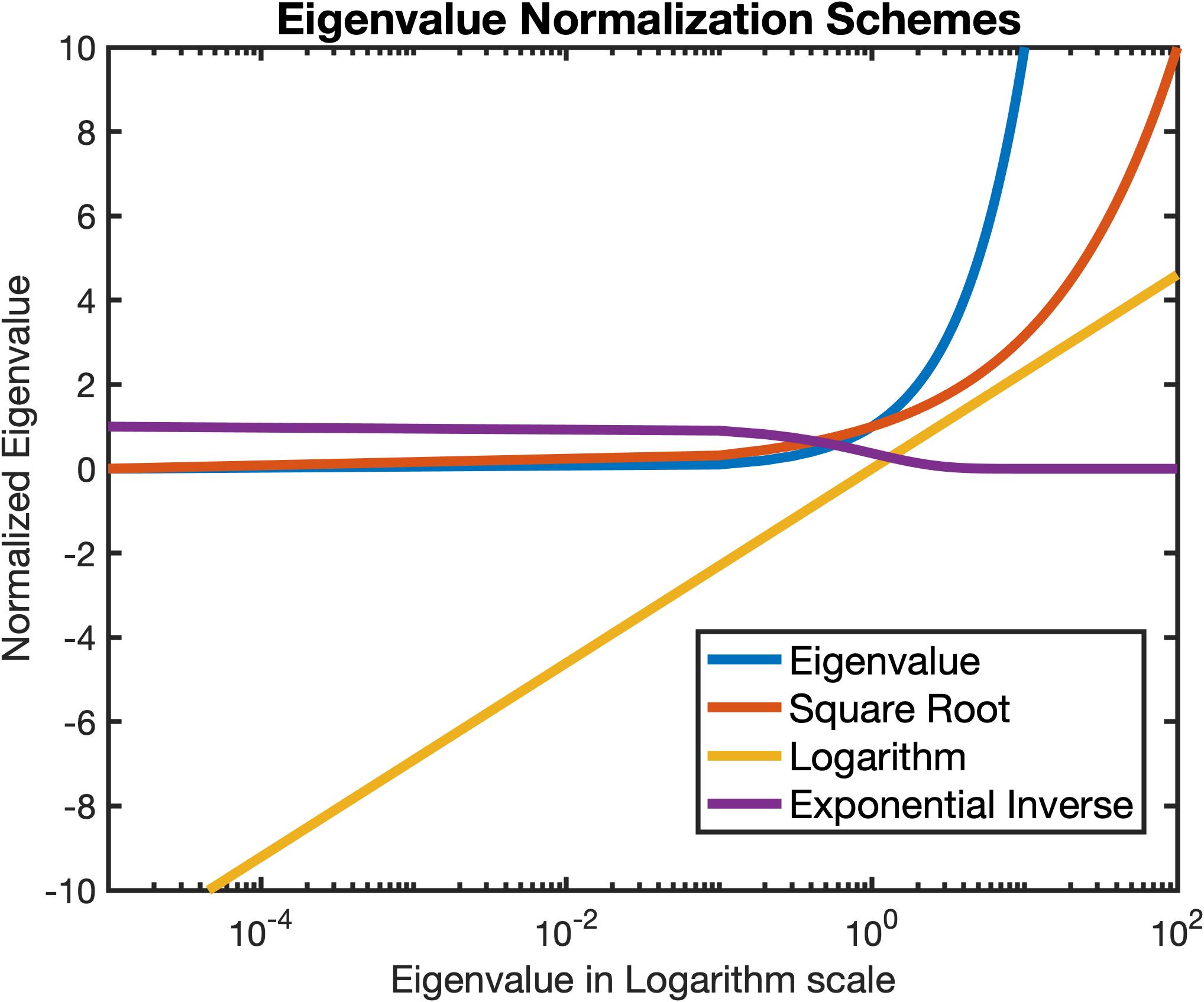}
    \captionsetup{font={small}}
    \caption{Different eigenvalue normalization schemes. Our proposed matrix exponential inverse can effectively narrow the eigenvalue range and amplify the importance of the small ones.}
    \label{fig:semilog_eigen}
\end{figure}

\subsection{Dynamic Scaling Factor} 
\label{sec:scaling_factor}

To scale the covariance square root $\mathbf{Q}$, one may consider multiplying a constant $a$: 
\begin{equation}
    a\cdot\mathbf{Q} = \mathbf{U}\{a {\lambda_{1}^{\frac{1}{2}}},\dots,a {\lambda_{d}^{\frac{1}{2}}}\}_{\rm diag}\mathbf{U}^{T}
\end{equation}
where $\{\cdot\}_{\rm diag}$ means transforming a vector to a diagonal matrix. It is non-trivial to choose a suitable constant $a$, as each covariance matrix has a different eigenvalue distribution. The scaling factor should accordingly be different and preferably dependent on the eigenvalues. An intuitive choice is using $||\mathbf{Q}||_{\rm F}$ or $||\mathbf{P}||_{\rm F}$ as the factor. However, both $\mathbf{Q}$ and $\mathbf{P}$ have very imbalanced eigenvalue distributions and are likely to be ill-conditioned (\emph{i.e.,} $\frac{\lambda_{1}}{\lambda_{d}}$ is too large). In that case, the small eigenvalues will contribute little and the first eigenvalue $\lambda_{1}$ will dominate the scaling factor (\emph{i.e.,} $||\mathbf{P}||_{\rm F}{\approx}\lambda_{1}$ and $||\mathbf{Q}||_{\rm F}{\approx}\lambda_{1}^{\frac{1}{2}}$). To avoid this issue, we propose to first balance the eigenvalue distribution by calculating the exponential inverse of the covariance:
\begin{equation}
    \mathbf{S}{=}e^{-\mathbf{P}} {=}\mathbf{U}e^{-\mathbf{\Lambda}}\mathbf{U}^{T}{=}\mathbf{U}\{e^{-\lambda_{1}},\dots,e^{-\lambda_{d}}\}_{\rm diag}\mathbf{U}^{T}
\end{equation}
Unfortunately, differentiating this step is not supported by the AutoGrad package of the deep learning frameworks. The gradient has to be manually derived. Given the partial derivatives of the loss $l$ w.r.t $\mathbf{S}$, we have ${\rm d}\mathbf{S}{=}{\rm d}\mathbf{U}e^{-\mathbf{\Lambda}}\mathbf{U}^{T}{+}\mathbf{U}{\rm d} e^{-\mathbf{\Lambda}}\mathbf{U}^{T}{+}\mathbf{U}e^{-\mathbf{\Lambda}}{\rm d}\mathbf{U}^{T}$ and ${\rm d}e^{-\mathbf{\Lambda}} {=} {-}\{e^{-\lambda_{1}}{,}{\dots}{,}e^{-\lambda_{d}}\}_{\rm diag}{\rm d}\mathbf{\Lambda}$. After some arrangements using the chain rule and matrix back-propagation rule~\cite{ionescu2015training}, the partial derivatives of the loss $l$ w.r.t the eigenvalue and eigenvector are calculated as:
\begin{equation}
    \begin{gathered}
    \frac{\partial l}{\partial \mathbf{U}}{=}\Big(\frac{\partial l}{\partial \mathbf{S}} {+} (\frac{\partial l}{\partial \mathbf{S}})^{T}\Big)\mathbf{U}\{e^{-\lambda_{1}},\dots,e^{-\lambda_{d}}\}_{\rm diag},\\
    \frac{\partial l}{\partial \mathbf{\Lambda}}{=}{-}\Big(\{e^{-\lambda_{1}},\dots,e^{-\lambda_{d}}\}_{\rm diag}\mathbf{U}^{T} \frac{\partial \it{l}}{\partial \mathbf{S}} \mathbf{U}\Big)_{\rm{diag}}
    \end{gathered}
    \label{inv_exp_de}
\end{equation}
where $(\cdot)_{\rm diag}$ denotes the operation of setting off-diagonal elements to zero. The covariance exponential inverse $\mathbf{S}$ normalizes the eigenvalue distribution on $(0,1]$ but reverses the order of importance (see also Fig.~\ref{fig:semilog_eigen}). We then combine $\mathbf{S}$ and $\mathbf{Q}$ to compute their cross-covariance:
\begin{equation}
    {\mathbf{Q}\mathbf{S}^{\it T}}=\mathbf{U} \{{\lambda_{1}^{\frac{1}{2}}e^{-\lambda_{1}}},\dots,{\lambda_{d}^{\frac{1}{2}}e^{-\lambda_{d}}}\}_{\rm diag}\mathbf{U}^{T}
    \label{eq:compensation}
\end{equation}
Eq.~\eqref{eq:compensation} actually measures the similarity between $\mathbf{Q}$ and $\mathbf{S}$. Compared with the ordinary covariance square root $\mathbf{Q}$, it has a more balanced eigenvalue distribution. Thus we use its Frobenius norm $||\mathbf{Q}\mathbf{S}||_{\rm F}$ as the scaling factor to compensate for the covariance square root.

\subsection{Covariance Square Root Compensation}

After generating the scaling factor, we multiply it to the covariance square root plus the ordinary one. This process can be denoted by:
\begin{equation}
    \mathbf{A} = ||\mathbf{Q}\mathbf{S}^{T}||_{\rm F}\cdot\mathbf{Q} + \mathbf{Q} = (||\mathbf{Q}\mathbf{S}^{T}||_{\rm F}+1)\cdot\mathbf{Q}
\end{equation}
where $\mathbf{A}$ is the final representation that will be fed to the fully-connected layer. The value of the Frobenius norm $||\mathbf{Q}\mathbf{S}^{T}||_{\rm F}$ takes the following form:
\begin{equation}
    ||\mathbf{Q}\mathbf{S}^{\it T}||_{\rm F} = \sqrt{{\rm Tr}((\mathbf{Q}\mathbf{S}^{\it T})(\mathbf{Q}\mathbf{S}^{\it T})^{H})} = \sqrt{\sum_{i=1}^{d}\Big(\lambda_{i}^{\frac{1}{2}}e^{-\lambda_{i}}\Big)^2}\\
    \label{eq:f_norm}
\end{equation}

\begin{table*}[htbp]
\captionsetup{font={small}}
\caption{Comparison with the state-of-the-art GCP approaches. For a fair comparison, we report our training results of MPN-COV~\cite{li2017second} and iSQRT-COV~\cite{li2018towards} using the same deep learning platform (\textsc{Pytorch}). Notice that the values reported in these two papers are not based on the commonly used backbones with the first-order pooling layers (see Supplementary Material for details). The values of other methods are taken directly from their papers as they are less related to our approach.} 
    \centering
    \resizebox{0.9\linewidth}{!}{
    \label{tab:performances}
    \begin{tabular}{r|c|c|c|c}
    \toprule
        Backbone & Method & Birds~\cite{WelinderEtal2010} & Aircrafts~\cite{maji2013fine} & Cars~\cite{KrauseStarkDengFei-Fei_3DRR2013}  \\
        \hline
        \multirow{10}*{VGG-16~\cite{simonyan2014very}} & B-CNN$_{2015}$~\cite{lin2015bilinear}  & 84.1 & 86.6 &91.3\\
        & Improved B-CNN$_{2017}$~\cite{lin2017improved}  &85.8 &88.5 & 92.0\\
        & LRBP$_{2017}$~\cite{Kong_2017_CVPR}  &84.2 &87.3 & 90.9\\
        & KP$_{2017}$~\cite{Cui_2017_CVPR}  & 86.2 & 86.9 & 92.4\\
        & MoNet$_{2018}$~\cite{gou2018monet} & \textbf{86.4} & 89.3 & 91.8\\
        & GP$_{2018}$~\cite{Wei_2018_ECCV}  &85.8 & 89.8 &92.8\\
        & CBP~\cite{gao2016compact} + RUN$_{2020}$~\cite{yu2020toward} & 85.7 & \textbf{91.0} &-- \\
        & iSQRT-COV$_{2018}$~\cite{li2018towards} &83.1 &88.6 &90.1\\
        & MPN-COV$_{2017}$~\cite{li2017second} & 83.4 &89.2 &90.9\\
        & MPN-COV + Our SEB & 85.7 ($\uparrow$ 2.3) & 90.7 ($\uparrow$ 1.5) & \textbf{93.0} ($\uparrow$ 2.1) \\
        \hline
        \multirow{5}*{ResNet-50~\cite{he2016deep}} & CBP~\cite{gao2016compact} & 81.6 & 81.6 & 88.6\\
        &KP~\cite{Cui_2017_CVPR} & 84.7 & 85.7 & 91.1 \\
        &iSQRT-COV$_{2018}$~\cite{li2018towards} &84.7 &89.6 &91.4 \\
        &MPN-COV$_{2017}$~\cite{li2017second} &84.3 &89.9 &91.7 \\
        &MPN-COV + Our SEB & \textbf{86.2} ($\uparrow$ 1.9) & \textbf{91.4} ($\uparrow$ 1.5) & \textbf{93.6}  ($\uparrow$ 1.9)\\
        \hline
        \multirow{3}*{ResNet-101~\cite{he2016deep}} & 
        iSQRT-COV$_{2018}$~\cite{li2018towards}  &86.0 &90.3 & 91.5 \\
        &MPN-COV$_{2017}$~\cite{li2017second}  &85.7 &89.8 &91.7 \\
        &MPN-COV + Our SEB & \textbf{87.0} ($\uparrow$ 1.3) & \textbf{91.9} ($\uparrow$ 2.1) & \textbf{93.9} ($\uparrow$ 1.8)\\
        \hline
        \multirow{3}*{ResNet-152~\cite{he2016deep}} &iSQRT-COV$_{2018}$~\cite{li2018towards}  &85.9 &90.4 &92.2 \\
        &MPN-COV$_{2017}$~\cite{li2017second}  &86.1 &91.3 &92.5 \\
        &MPN-COV + Our SEB & \textbf{87.2} ($\uparrow$ 1.1) & \textbf{92.7} ($\uparrow$ 1.4) & \textbf{94.1} ($\uparrow$ 1.6)\\
        \hline
        \multirow{3}*{EfficientNet-b5~\cite{tan2019efficientnet}}
        &iSQRT-COV$_{2018}$~\cite{li2018towards}  &87.1 &92.8 & 93.3\\
        &MPN-COV$_{2017}$~\cite{li2017second}  & 87.3 &92.4 &93.4 \\
        &MPN-COV + Our SEB & \textbf{88.2} ($\uparrow$ 0.9) & \textbf{93.5} ($\uparrow$ 1.1) & \textbf{94.6} ($\uparrow$ 1.2)\\
        \hline
        Average Gain & Our SEB & $\uparrow$ 1.5 & $\uparrow$ 1.5 & $\uparrow$ 1.7 \\
    \bottomrule
    \end{tabular}
    }
\end{table*}

It is easy to find that the numerical value is in the range of $(0,||\mathbf{Q}||_{\rm F})$. We add the ordinary matrix square root to ensure that $\mathbf{A}$ can always amplify the eigenvalues of $\mathbf{Q}$ (\emph{i.e.,} $||\mathbf{Q}\mathbf{S}^{T}||_{\rm F}{+}1{>}1$). Note that the norm $||\mathbf{Q}\mathbf{S}^{T}||_{\rm F}$ is differentiable and the gradient could help to increase the significance of small eigenvalues. Its derivatives w.r.t. $\mathbf{Q}$ and $\mathbf{S}$ are given by:
\begin{equation}
    \begin{aligned}
    \frac{\partial ||\mathbf{Q}\mathbf{S}^{\it T}||_{\rm F}}{\partial \mathbf{Q}} &= \frac{1}{||\mathbf{Q}\mathbf{S}^{\it T}||_{\rm F}}\mathbf{Q}\mathbf{S}^{T}\mathbf{S}\\
    &=\mathbf{U} \{\frac{{\lambda_{1}^{\frac{1}{2}}e^{-2\lambda_{1}}}}{||\mathbf{Q}\mathbf{S}^{\it T}||_{\rm F}},\dots,\frac{{\lambda_{d}^{\frac{1}{2}}e^{-2\lambda_{d}}}}{||\mathbf{Q}\mathbf{S}^{\it T}||_{\rm F}}\}_{\rm diag}\mathbf{U}^{T},\\
    \frac{\partial ||\mathbf{Q}\mathbf{S}^{\it T}||_{\rm F}}{\partial \mathbf{S}} &= \frac{1}{||\mathbf{Q}\mathbf{S}^{\it T}||_{\rm F}}\mathbf{S}\mathbf{Q}^{T}\mathbf{Q}\\
    &=\mathbf{U} \{\frac{{\lambda_{1}e^{-\lambda_{1}}}}{||\mathbf{Q}\mathbf{S}^{\it T}||_{\rm F}},\dots,\frac{{\lambda_{d}e^{-\lambda_{d}}}}{||\mathbf{Q}\mathbf{S}^{\it T}||_{\rm F}}\}_{\rm diag}\mathbf{U}^{T}
    \end{aligned}
\end{equation}

For both matrices, the gradients are healthy as the use of $\mathbf{S}$ balances the eigenvalue distribution and reduces the risk of ill-conditioned derivative matrices. This could help the network to generate the better-conditioned covariance matrix $\mathbf{P}$ and the covariance square root $\mathbf{Q}$, which implicitly amplifies the importance of the small eigenvalues. 

\section{Experiments}
\label{sec:5}


\subsection{Implementation Details}

The code is implemented in \textsc{Pytorch}. We run all the experiment on the workstation equipped with a GeForce GTX 1080 Ti GPU and a 6-core Intel Core i7-7800X@3.50GHz CPU. For MPN-COV~\cite{li2017second} and our proposed SEB, the forward eigendecomposition is conducted on the CPU for a faster speed. The other operations are performed on the GPU.

\subsubsection{Dataset Specifications} The Birds dataset includes $11,788$ images belonging to $200$ bird species. The Aircrafts dataset contains $10,000$ images of $100$ classes of airplanes, and the Cars dataset consists of $16,185$ images from $196$ classes. Besides the three commonly used datasets, we also evaluate our method on two large datasets, namely Dogs and iNats. The Dogs dataset is comprised of $20,580$ images of $120$ dog categories, and the INats dataset have $675,170$ images from $5,089$ natural fine-grained categories that belong to $13$ super-categories.

\subsubsection{Perturbation-based Explainability Settings} The image is resized to $448{\times}448$ before being fed into the network. We keep the weights of the model frozen and only update the image pixel values. The learning rate to perturb the image is set to $0.1$ and the perturbation lasts for $1,000$ iterations for each image.

For the training recipe of different models on fine-grained benchmarks, please refer to Supplementary Material for details.




\subsection{Results Compared with GCP Methods}

Table~\ref{tab:performances} compares the validation accuracy on three benchmarks with different backbones against the standard GCP method (\emph{i.e.,} MPN-COV~\cite{li2017second}), as well as iSQRT-COV~\cite{li2018towards} and other covariance pooling methods. We can observe that the standard GCP method equipped with our proposed SEB improves the performance of its original version~\cite{li2017second} by $1.6$\% on average across different datasets regardless of the backbones. To be more specific, on Birds dataset, the performance is improved by $1.5$\% across backbones.
On Aircrafts dataset, our SEB improves MPN-COV~\cite{li2017second} by $1.5$\% on average. On Cars dataset, the average performance gain brought by SEB is $1.7$\%. Based on ResNet-152~\cite{he2016deep} and EfficientNet-b5~\cite{tan2019efficientnet}, our method achieves the state-of-the-art performance of GCP methods on all datasets.

The consistent performance gain demonstrates that our proposed SEB could be a powerful add-on for the GCP methods on the fine-grained recognition task. With the VGG backbone, our method is slightly inferior to RUN~\cite{yu2020toward} and MoNet~\cite{gou2018monet} on Aircrafts and Birds, respectively. That is mainly because the covariance of VGG architecture is of the size $512{\times}512$, which is twice larger than ResNet~\cite{he2016deep} and more prone to be ill-conditioned. However, there are two sides to reducing the number of channels. Projecting the channel dimension into $256$ could reduce the covariance size but might weaken the representation power and might not necessarily lead to performance improvements.
Besides, these methods do not report the experimental results on other backbones and their applicability on other deep architectures cannot be confirmed. It is also worth mentioning that our SEB does not introduce any additional parameters to MPN-COV~\cite{li2017second} and only slightly increases the speed and memory consumption (See Sec.~\ref{sec:speed_memory}). 

To sum up, the small eigenvalues play a vital role on the fine-grained classification task as they capture the semantic class-specific features, and amplifying their importance does bring consistent improvements. However, the small eigenvalues seem less important on the large-scale classification dataset (\emph{i.e.,} ImageNet~\cite{deng2009imagenet}) in which the inter-class feature differences are no longer subtle and are more likely to be encoded by the large eigenvalues. The small eigenvalues might only capture the data noise and the network predictions would be largely dependent on large eigenvalues. For the detailed analysis, please refer to Supplementary Material which illustrates this point by a series of experiments including the explainability visualizations.

\subsection{Results Compared with Other FGVC Approaches}

\begin{table}[htbp]
\caption{Comparison with other state-of-the-arts that are achieved by transformer-based methods on larger datasets.}
 \centering
    \resizebox{0.99\linewidth}{!}{
    \label{tab:extra_performances_sota}
    \setlength{\tabcolsep}{1pt}
    \begin{tabular}{r|c|c|c|c}
    \toprule
        Backbone & Method & Cars~\cite{KrauseStarkDengFei-Fei_3DRR2013} & Dogs~\cite{khosla2011novel} & INats~\cite{van2018inaturalist} \\
        \hline
        \multirow{3}*{ViT~\cite{dosovitskiy2020image}} &ViT$_{2021}$~\cite{dosovitskiy2020image} &93.5 &91.2 &68.0 \\ 
        &TransFG$_{2021}$~\cite{he2021transfg} &94.1 &92.3 &71.7 \\ 
        &AFTrans$_{2021}$~\cite{zhang2021free} &\textbf{95.0} &91.6 &68.9 \\ 
        \hline
        \multirow{3}*{EfficientNet-b5~\cite{tan2019efficientnet}}
        &iSQRT-COV~\cite{li2018towards}   &93.3 &92.3 &71.0\\
        &MPN-COV~\cite{li2017second}  &93.4 &92.1 &70.8\\
        &MPN-COV + Our SEB & {94.6} ($\uparrow$ 1.2) & \textbf{93.0} ($\uparrow$ 0.9) & \textbf{72.3} ($\uparrow$ 1.5) \\
    \bottomrule
    \end{tabular}
    }
\end{table}

Besides the evaluation within the scope of GCP methods, we also compare our method with other state-of-the-art FGVC approaches on some larger datasets. Table~\ref{tab:extra_performances_sota} presents the validation accuracy of our approach and recent transformer-based FGVC methods on the three larger datasets, \emph{i.e.,} Cars~\cite{KrauseStarkDengFei-Fei_3DRR2013}, Dogs~\cite{khosla2011novel}, and INats~\cite{van2018inaturalist}. Our method outperforms other baselines by $0.7\%$ on Dogs~\cite{khosla2011novel} and by $0.6\%$ on INats~\cite{van2018inaturalist}, while the performance slight falls behind TransFG~\cite{he2021transfg} by $0.4\%$ on Cars~\cite{KrauseStarkDengFei-Fei_3DRR2013}. This observation indicates that our method can also have very competitive performances against other FGVC approaches on large benchmarks.






\subsection{Results on Other Deep Models}

\begin{table}[htbp]
\caption{Performance on other popular deep architectures.}
 \centering
    \resizebox{0.99\linewidth}{!}{
    \label{tab:extra_performances}
    \setlength{\tabcolsep}{1pt}
    \begin{tabular}{r|c|c|c|c}
    \toprule
        Backbone & Method & Birds~\cite{WelinderEtal2010} & Aircrafts~\cite{maji2013fine} & Cars~\cite{KrauseStarkDengFei-Fei_3DRR2013}  \\
        \hline
        \multirow{3}*{Inception V3~\cite{szegedy2016rethinking}} &iSQRT-COV~\cite{li2018towards}  &83.5 &90.0 &90.5 \\
        &MPN-COV~\cite{li2017second}  &82.7 &89.9 &90.6 \\
        &MPN-COV + Our SEB & \textbf{84.8}  ($\uparrow$ 2.1) & \textbf{90.9} ($\uparrow$ 1.0) & \textbf{92.5} ($\uparrow$ 1.9)\\
        \hline
        \multirow{3}*{DenseNet-169~\cite{huang2017densely}} &iSQRT-COV~\cite{li2018towards}  &86.4 &91.1 &92.6 \\
        &MPN-COV~\cite{li2017second}  &86.3 &90.8 & 92.9 \\
        &MPN-COV + Our SEB &  \textbf{87.6} ($\uparrow$ 1.3) &\textbf{92.3}  ($\uparrow$ 1.5) & \textbf{94.0} ($\uparrow$ 1.1)\\
    \bottomrule
    \end{tabular}
    }
\end{table}

Under the same training protocol of ResNet, we conduct more experiments on other popular deep architectures, namely Inception~\cite{szegedy2016rethinking} and DenseNet~\cite{huang2017densely}. Table~\ref{tab:extra_performances} displays the performances on the fine-grained benchmarks. Similar with the results on VGG and ResNet, our proposed SEB significantly improves MPN-COV~\cite{li2017second} by $1.5$\% on average. This consistent performance gain demonstrates the general model-agnostic applicability of our SEB for different deep architectures.

\subsection{Ablation Studies}

Using ResNet-50 as the backbone, we conduct two ablation studies on the impact of using only SEB path and the impact of different scaling factors.

\begin{figure*}[h]
    \centering
    \includegraphics[width=0.8\linewidth]{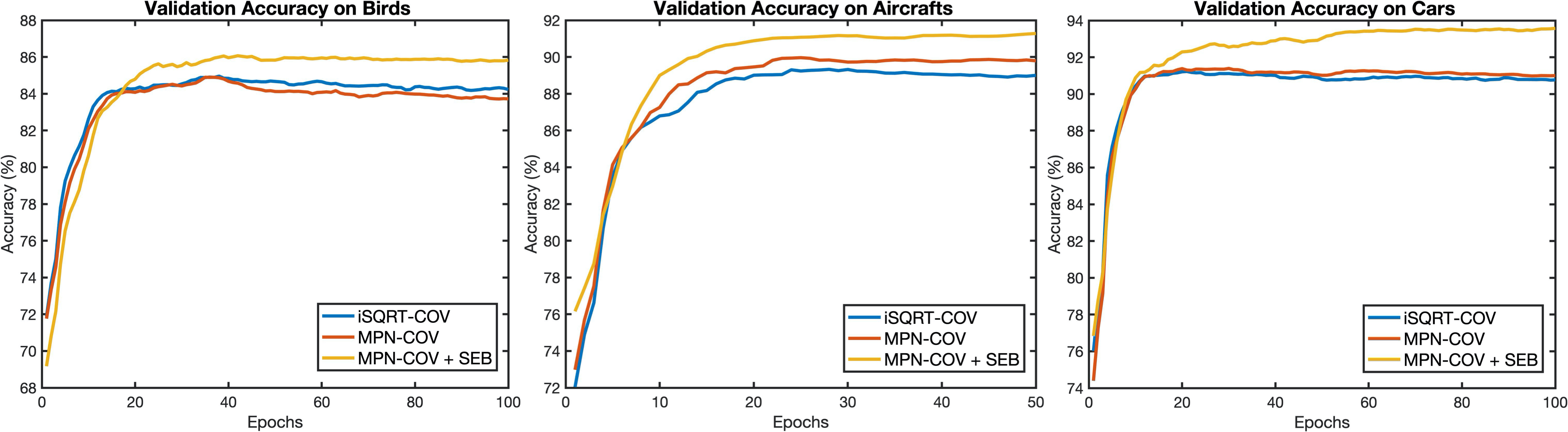}
    \caption{Validation accuracy curves versus training epochs of ResNet-50 on three fine-grained benchmarks. Our proposed SEB consistently outperforms the original GCP methods by a large margin. The lines are smoothed by a moving average filter for a better view.}
    \label{fig:fgvc_val}
\end{figure*}

\begin{figure*}[h]
    \centering
    \includegraphics[width=0.9\linewidth]{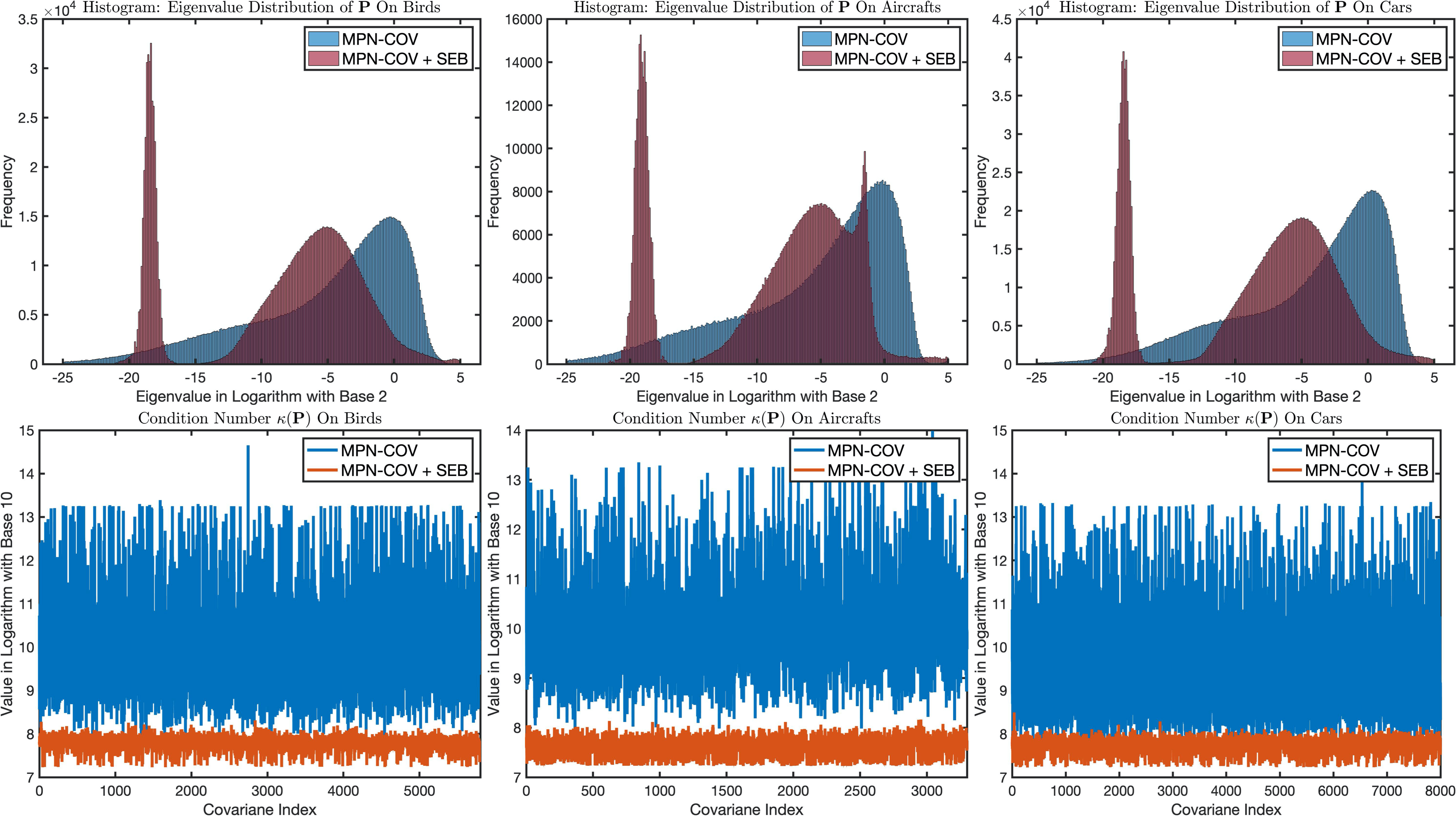}
    \caption{(\emph{Top}) The histogram of eigenvalue distribution of the covariance $\mathbf{P}$ of MPN-COV~\cite{li2017second} and our SEB on three fine-grained benchmarks. The proposed SEB effectively narrows the eigenvalue range and magnify the small ones. (\emph{Bottom}) The condition number $\kappa(\mathbf{P})$ of MPN-COV~\cite{li2017second} and our proposed SEB on the fine-grained benchmarks. Our covariance matrices are consistently better-conditioned than MPN-COV~\cite{li2017second} and the relative importance of small eigenvalues are amplified.}
    \label{fig:eigen_p}
\end{figure*}

\subsubsection{Impact of Only SEB Path.} 

\begin{table}[htbp]
\setlength{\tabcolsep}{1pt}
\centering
\captionsetup{font={small}}
\caption{Impact of using only SEB path.}\label{tab:path}
\resizebox{0.9\linewidth}{!}{
\begin{tabular}{ccccc}\toprule  
Baselines & Formulation &Birds~\cite{WelinderEtal2010} & Aircrafts~\cite{maji2013fine} & Cars~\cite{KrauseStarkDengFei-Fei_3DRR2013} \\ \hline
MPN-COV~\cite{li2017second} & $\mathbf{Q}$ &84.3 &89.9 &91.7 \\ \hline
SEB & $||\mathbf{Q}\mathbf{S}^{\it T}||_{\rm F}{\cdot}\mathbf{Q}$ &85.6 &91.1 &93.0 \\  \hline
MPN-COV+SEB & $||\mathbf{Q}\mathbf{S}^{\it T}||_{\rm F}{\cdot}\mathbf{Q}{+}\mathbf{Q}$ &\textbf{86.2} & \textbf{91.4} & \textbf{93.6}\\  \bottomrule
\end{tabular}
}
\end{table}

Instead of combining the matrix square root and the compensated one $||\mathbf{Q}\mathbf{S}^{\it T}||_{\rm F}{\cdot}\mathbf{Q}{+}\mathbf{Q}$, one may consider only passing the amplified covariance square root $||\mathbf{Q}\mathbf{S}^{\it T}||_{\rm F}{\cdot}\mathbf{Q}$ to the fully-connected layer. Table~\ref{tab:path} compares the performances of using different paths. Solely using the SEB path $||\mathbf{Q}\mathbf{S}^{\it T}||_{\rm F}{\cdot}\mathbf{Q}$ brings about $1.2$\% increase on the performances over MPN-COV~\cite{li2017second}. The combination of the two paths can assure that the final representation always amplifies the covariance square root $\mathbf{Q}$, thus outperforming using only SEB by $0.4$\% on average. 

\subsubsection{Impact of Scaling Factors.}

\begin{table}[htbp]
\centering
\captionsetup{font={small}}
\caption{Impact of different scaling factors.}\label{tab:normalization}
\resizebox{0.95\linewidth}{!}{
\begin{tabular}{cccc}\toprule  
Factor & Birds~\cite{WelinderEtal2010} & Aircrafts~\cite{maji2013fine} & Cars~\cite{KrauseStarkDengFei-Fei_3DRR2013} \\\hline
Constant 100 &82.4 &88.5 &89.3\\ \hline
$\texttt{ReLU}(\texttt{MLP} (\mathbf{\Lambda^{\frac{1}{2}}}))+1$ &82.9 &89.1 &91.3 \\ \hline
$\texttt{ReLU}(\texttt{MLP} (\mathbf{Q}))+1$  &80.5 &88.3 &90.7 \\ \hline
$||\mathbf{S}||_{\rm F}$ &83.3 &90.1 &90.6\\  \hline
$||\mathbf{Q}||_{\rm F}$ &83.8 &90.3 &92.1 \\  \hline
$||\mathbf{Q}\mathbf{S}^{T}||_{\rm F}$ &\textbf{86.2} & \textbf{91.4} & \textbf{93.6}\\  \bottomrule
\end{tabular}
}
\end{table}

To magnify the eigenvalues of $\mathbf{Q}$, one may think about using other scaling factors instead of our used $||\mathbf{Q}\mathbf{S}^{T}||_{\rm F}$. We take some possible factors to evaluate the impact, including the fixed constant, $||\mathbf{Q}||_{\rm F}$, $||\mathbf{S}||_{\rm F}$, and factors learned by an MLP from $\mathbf{\Lambda}^{\frac{1}{2}}$ and $\mathbf{Q}$. As shown in Table~\ref{tab:normalization}, none of the alternative factors bring obvious improvements and some even have side effects on the performances. We conjecture that this is related to the backward gradient of the scaling factor. Using a constant factor and $||\mathbf{S}||_{\rm F}$ cannot propagate valid gradients to $\mathbf{Q}$. For the scaling factor $||\mathbf{Q}||_{\rm F}$, its derivative w.r.t. $\mathbf{Q}$ is $\mathbf{Q}{/}{||\mathbf{Q}||_{\rm F}}$, which is prone to be ill-conditioned. For the learned scaling factor, the value is unbounded and the gradient can not be constrained in good conditions. By contrast, our $||\mathbf{Q}\mathbf{S}^{T}||_{\rm F}$ consistently has well-conditioned gradient matrices and outperforms other schemes significantly. 

\subsection{Speed and Memory Comparison}
\label{sec:speed_memory}

Table~\ref{tab:speed} compares the time and memory consumption of the GCP models for each mini-batch. Compared with the ordinary MPN-COV~\cite{li2017second}, our proposed SEB only brings marginal overhead on the computation complexity and memory cost. The extra memory usage is due to the calculation and storage of the scaling factor in the proposed SEB path. 

\begin{table}[htbp]
    \centering
    \caption{Time and memory consumption of GCP methods with ResNet-50 model on Cars dataset for each mini-batch. 
    }\label{tab:speed}
    \resizebox{0.99\linewidth}{!}{
    \begin{tabular}{r|c|c}
    \toprule
    Methods  & Computational Time (s) & Memory Usage (M)\\
    \hline
    iSQRT-COV~\cite{li2018towards} &0.28 &4345\\
    MPN-COV~\cite{li2017second} &0.30 &4298\\
   MPN-COV + Our SEB & 0.31 & 4501\\
    \bottomrule
    \end{tabular}
    }
\end{table}

\subsection{Validation Accuracy Curves}

Fig.~\ref{fig:fgvc_val} displays the validation accuracy curves of our SEB and the GCP methods on three fine-grained benchmarks. As can be observed, our proposed SEB consistently brings the performance gain about $2\%$ over MPN-COV~\cite{li2017second} and iSQRT-COV~\cite{li2018towards} during the training process. Besides, the original GCP methods show some signs of overfitting, \emph{i.e.,} their validation accuracy seems to slowly decrease as the training goes in the later stage. In contrast, our SEB appears to continuously improve throughout the training.

\subsection{Eigenvalue Distribution Comparison}
In Sec.~\ref{sec:scaling_factor}, we show that the backward gradient of $||\mathbf{Q}\mathbf{S}^{T}||_{\rm F}$ are healthy and can help the network to generate better-conditioned covariance matrices. To validate the concrete impact, we compare the eigenvalue distributions of our proposed SEB method and the ordinary GCP method MPN-COV~\cite{li2017second}. The comparison is made by visualizing the histogram of the eigenvalues and by evaluating the matrix condition number. The matrix condition number $\kappa(\cdot)$ is defined as:
\begin{equation}
    \kappa(\mathbf{X})={||\mathbf{X}||_{2}}{||\mathbf{X}^{-1}||_{2}}=\frac{\sigma_{max}(\mathbf{X})}{\sigma_{min}(\mathbf{X})}
\end{equation}
where $\mathbf{X}$ is the matrix to be measured. The condition number computes the ratio between the maximum and minimum eigenvalues. The lower the ratio is, the higher the relative importance of small eigenvalues will be. The histogram displays the overall eigenvalue distribution for all covariance matrices, whereas the condition number measures the conditioning of each individual matrix. The joint use of these two evaluation methods can provide a comprehensive picture of the eigenvalue information. 


Fig.~\ref{fig:eigen_p} top shows the histogram of eigenvalues distribution for the global covariance $\mathbf{P}$. As can be observed, our SEB effectively narrows the eigenvalue range. The large eigenvalues are reduced and the small ones are increased. Interestingly, on all datasets, a large portion of small eigenvalues group together to form a peak around $2^{-18}$. We expect that these eigenvalues and associated eigenvectors are the determinant factors that capture the rich semantic information and boost the performances of the GCP methods. The condition number of all the covariance matrices $\kappa(\mathbf{P})$ are plotted in the bottom of Fig.~\ref{fig:eigen_p}. For MPN-COV~\cite{li2017second}, the value varies from $10^{8}$ to $10^{15}$, whereas the condition number of our SEB is on $[10^7,10^8]$. This demonstrates that our covariance matrices are consistently much better-conditioned than MPN-COV~\cite{li2017second}.

\section{Conclusion}
\label{sec:6}

In this paper, we propose two explainability methods to diagnose the eigenvalue behaviors of GCP layers. Both quantitative evaluation and visual observation demonstrate that the small eigenvalues capture more class-relevant and semantically meaningful feature patterns. Based on this finding, we further propose a plug-in network branch to magnify the importance of small eigenvalues and the associated eigenvectors. Without bringing extra parameters, the existing GCP method integrated with our proposed component achieves state-of-the-art performances of GCP methods on fine-grained benchmarks. Moreover, on larger datasets, our method also has competitive performances against other FGVC approaches.


%



\ifCLASSOPTIONcompsoc
\fi


\ifCLASSOPTIONcaptionsoff
  \newpage
\fi



\bibliographystyle{IEEEtran}
\bibliography{egbib}
%


%

\begin{IEEEbiography}[{\includegraphics[width=1in,height=1.25in,keepaspectratio]{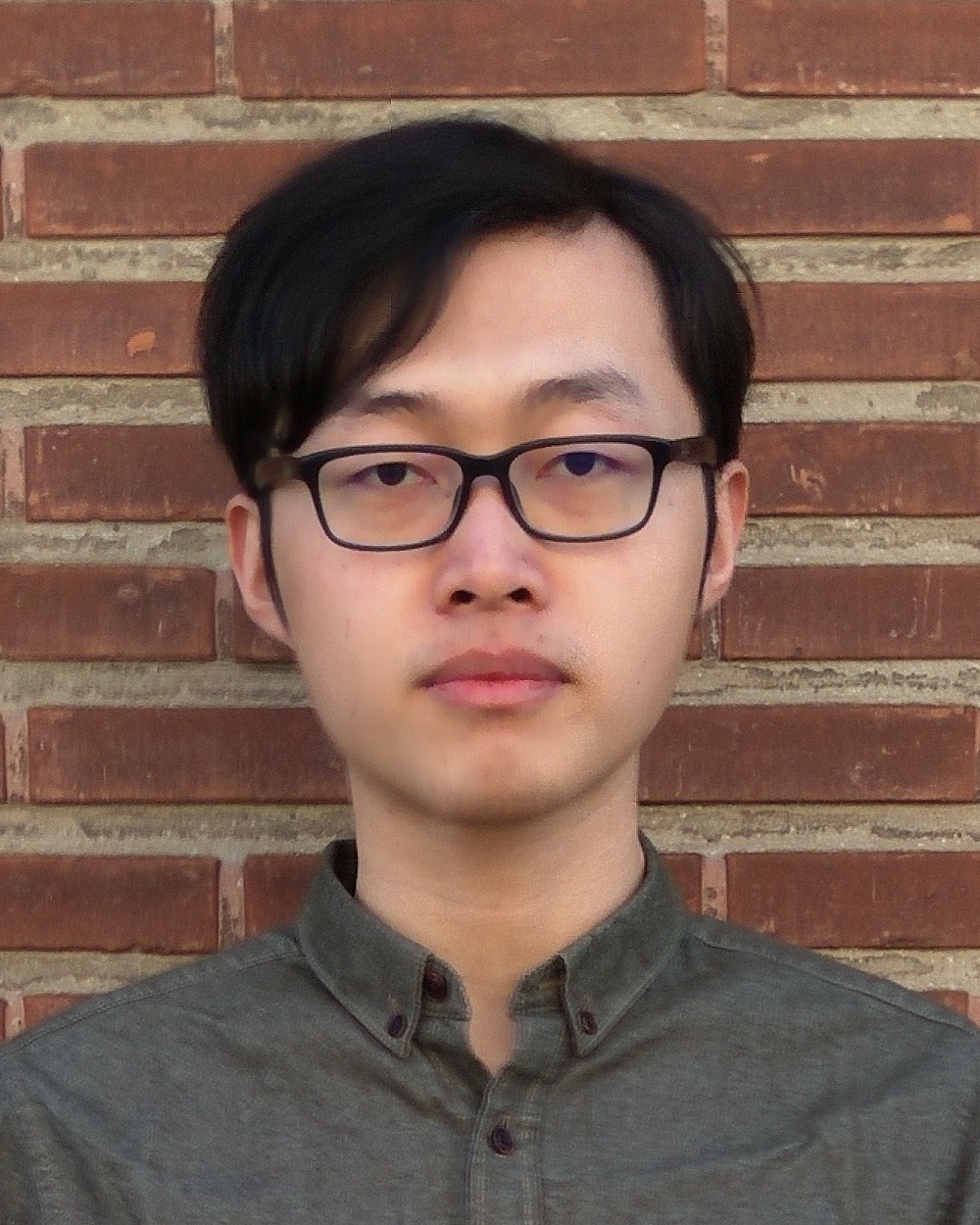}}]{Yue Song}
received the B.Sc. \emph{cum laude} from KU Leuven, Belgium and the joint M.Sc. \emph{summa cum laude} from the University of Trento, Italy and KTH Royal Institute of Technology, Sweden. Currently, he is a Ph.D. student with the Multimedia and Human Understanding Group (MHUG) at the University of Trento, Italy. His research interests are computer vision and numerical methods.
\end{IEEEbiography}

\begin{IEEEbiography}[{\includegraphics[width=1in,height=1.25in,keepaspectratio]{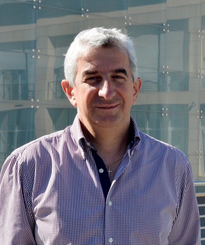}}]{Nicu Sebe} is Professor with the University of
Trento, Italy, leading the research in the areas of
multimedia information retrieval and human behavior
understanding. He was the General Co-
Chair of ACM Multimedia 2013, and the Program
Chair of ACM Multimedia 2007 and 2011, ECCV
2016, ICCV 2017 and ICPR 2020. He is a fellow
of the International Association for Pattern
Recognition.
\end{IEEEbiography}


\begin{IEEEbiography}[{\includegraphics[width=1in,height=1.25in,keepaspectratio]{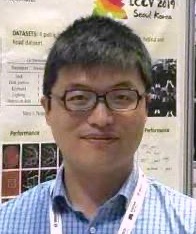}}]{Wei Wang}
is an Assistant Professor of Computer Science at University of Trento, Italy. Previously, after obtaining his PhD from University of
Trento in 2018, he became a Postdoc at EPFL,
Switzerland. His research interests include machine learning and its application to computer
vision and multimedia analysis.
\end{IEEEbiography}



\clearpage 
\appendices

\section{Implementation Details and Extra Visualization Results}

\subsection{Training Recipe for Fine-grained Benchmarks}
The ResNet and VGG pre-trained on ImageNet~\cite{deng2009imagenet} are fine-tuned on each fine-grained dataset. The images are first resized to the resolution of $512{\times}512$, then centered cropped to $448{\times}448$, and finally fed into the network. The $1000$-$d$ fully-connected layer of the original network is changed to fit the number of classes. The model is trained using SGD with momentum $0.9$. The batch size is set as $8$ for Aircrafts~\cite{maji2013fine} and set as $10$ for Birds~\cite{WelinderEtal2010} and Cars~\cite{KrauseStarkDengFei-Fei_3DRR2013}. We make the inference also on the $448{\times}448$ centered crop of the test image. Since the covariance matrix is symmetric, we only pass the upper triangular part to the fully-connected layer.

\subsubsection{ResNet Settings} 
For the ResNet architectures, we squeeze the channels of the final convolutional feature from $2048$ to $256$ for computation efficiency of covariance matrices. The resultant spatial dimension of the covariance is $256\times256$. The initial learning rate is set as $3{\times}10^{-3}$ for the convolutional layers. The learning rate of the fully-connected layer is set $5, 10,$ and $20$ times larger than the convolutional layers for Aircrafts~\cite{maji2013fine}, Cars~\cite{KrauseStarkDengFei-Fei_3DRR2013}, and Birds~\cite{WelinderEtal2010} respectively. The weight decay of the optimizer is set as $1{\times}10^{-3}$ for Aircrafts~\cite{maji2013fine} and as $1{\times}10^{-4}$ for Birds~\cite{WelinderEtal2010} and Cars~\cite{KrauseStarkDengFei-Fei_3DRR2013}. For Aircrafts~\cite{maji2013fine} dataset, the training lasts for $50$ epochs with dividing the learning rate by $10$ at epoch $20$. For Cars~\cite{KrauseStarkDengFei-Fei_3DRR2013} and Birds~\cite{WelinderEtal2010} datasets, the training lasts for $100$ epochs and the learning rate dividing happens at epoch $50$.

\subsubsection{VGG Settings} 
For the VGG architecture, the final representation has $512$ channels, which is much smaller than ResNet and will lead to the covariance matrix of a acceptable size. Therefore, we do not perform any channel reduction and the resultant covariance matrix is of size $512\times512$. For Cars~\cite{KrauseStarkDengFei-Fei_3DRR2013} dataset, the initial learning rate is set as $6{\times}10^{-3}$ for the fully-connected layer and $1.2{\times}10^{-3}$ for the other layers. We set the initial learning rate to $3{\times}10^{-3}$ for Aircrafts~\cite{WelinderEtal2010} and Birds~\cite{WelinderEtal2010}. Batch normalization is used to stabilize the training process. The training lasts $150$ epochs for all the datasets with periodically learning rate decay by $10$ at epoch $50$ and epoch $100$. 


For the concern of comparison fairness, the same training protocol is applied to  MPN-COV~\cite{li2017second} and iSQRT-COV~\cite{li2018towards} methods.



\subsection{Random Backward-based Visualization}

\begin{figure*}
    \centering
    \includegraphics[width=0.99\textwidth]{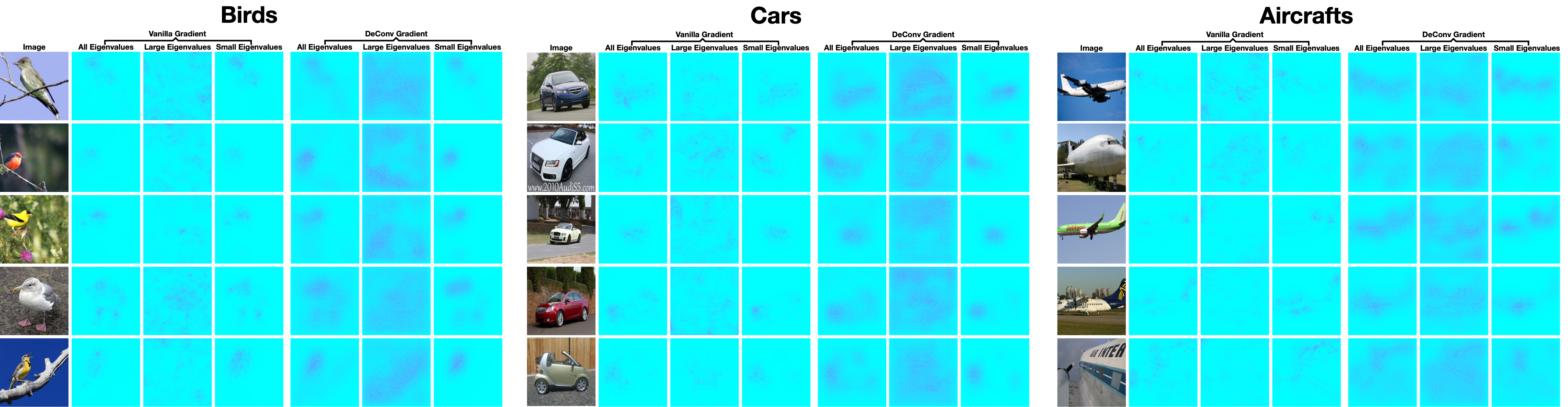}
    \caption{Visualization of randomly chosen backward-based attributions on three fine-grained datasets. The small eigenvalues generate the input activations that are more coherent with those using all the eigenvalues. Zoom in for a better view.}
    \label{fig:app_backward}
\end{figure*}

Fig.~\ref{fig:app_backward} displays randomly selected backward-based visualizations on the three fine-grained benchmarks. For all the datasets, the small eigenvalues display much more similar input activations with those of all the eigenvalues. This implies that the classification decisions of the GCP networks are more dependent on the small eigenvalues and the associated eigenvectors.

\subsection{Random Perturbation-based Visualization}

\begin{figure*}
    \centering
    \includegraphics[width=0.99\textwidth]{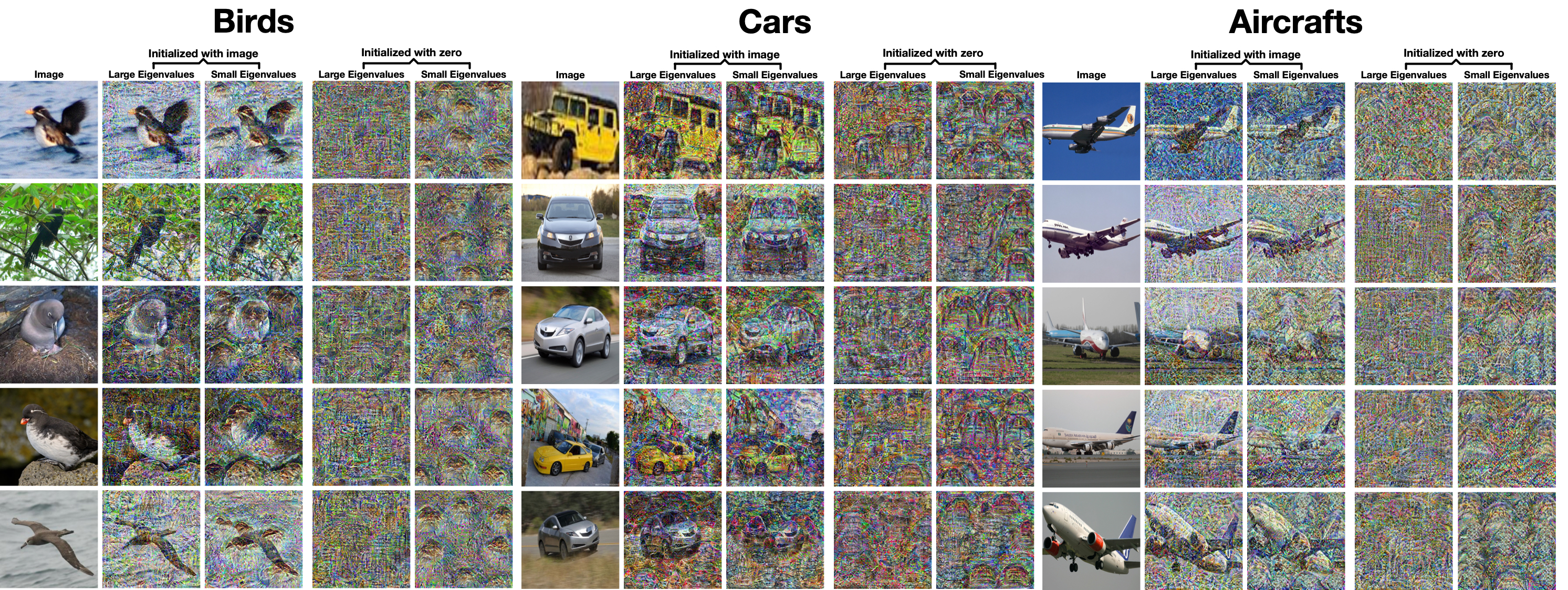}
    \caption{Randomly selected perturbation-based visualizations. The small eigenvalues mainly activates the class-relevant semantic feature, whereas the large eigenvalues are related to the background and the human-uninterpretable patterns. Zoom in for a better view.}
    \label{fig:app_perturb}
\end{figure*}

\begin{figure*}
    \centering
    \includegraphics[width=0.99\linewidth]{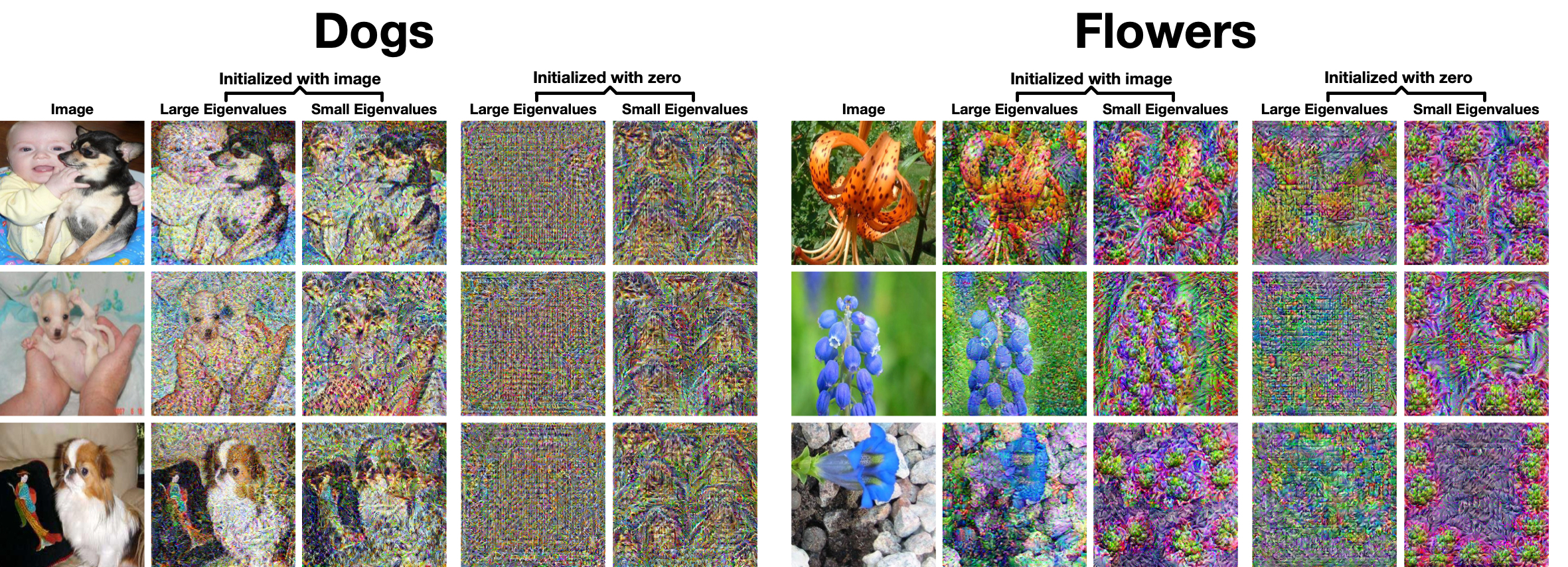}
    \caption{Extra visualizations on Dogs and Flowers datasets. Obviously, small eigenvalues capture more semantically-meaningful and class-relevant feature patterns.}
    \label{fig:app_perturb_dog_flower}
\end{figure*}

Fig.~\ref{fig:app_perturb} shows the randomly selected perturbation-based visualization. For Birds datatset, the visualization of small eigenvalues are interpretable, where the head, beak, and feather of the bird species emerge repeatedly in the images. However, the images on Cars and Aircrafts are less explainable. It might be difficult to tell which visualization characterizes more class-relevant feature patterns. We believe this is largely because the subtle differences between cars or airplanes are not easily interpretable by human beings. For example, classes of Cars datasets are typically at the level of Make, Model, and Year. One can not really tell the differences between the classes 'Audi S4 Sedan 2012' and 'Audi S4 Sedan 2007'. Thus the visualized feature patterns can be vague. The fine-grained datasets of natural species, on the other hand, might generate more human-friendly feature patterns. Fig.~\ref{fig:app_perturb_dog_flower} displays the perturbation-based visualizations of the ResNet-50 trained on Dogs~\cite{KhoslaYaoJayadevaprakashFeiFei_FGVC2011} and Flowers~\cite{Nilsback09a} dataset. Similar with the images on Birds, the visualization is interpretable and the feature patterns of small eigenvalues are more class-relevant. 

All in all, for the fine-grained datasets we have tested, the visualizations of the small eigenvalues generally have more semantically meaningful and more structured patterns. The large eigenvalues mainly generate not obviously class-relevant and hard-to-interpret features. 

\section{Experiments on ImageNet}
Since some GCP methods (\emph{i.e.,} MPN-COV~\cite{li2017second} and iSQRT-COV~\cite{li2018towards}) can work on the generic visual recognition task (\emph{e.g.}, ImageNet), we also apply our proposed SEB on this dataset to evaluate the performances. 

\subsection{Training Recipe for ImageNet}
We take AlexNet and ResNet-50 as the model architectures and use the same experimental settings in our conference paper~\cite{song2021approximate}. We set the batch size as $128$ for AlexNet and $256$ for ResNet. The AlexNet is trained for $30$ epochs with an initial learning rate set as $10^{-1.1}$. The learning rate decays by $10$ every $10$ epochs. For training ResNet, we use the same learning rate to train for $60$ epochs but decays by $10$ at epoch $30$ and epoch $45$. We use SGD for optimization, with momentum of $0.9$ and weight decay of $0.0001$ for ResNet and $0.0005$ for AlexNet. The network parameters are randomly initialized for both architectures. During training, the images are resized to $256{\times}256$ and then cropped to $224{\times}224$, with random horizontal flip augmentation. The inference is conducted on the $224{\times}224$ centered crop from the validation set. 

\subsection{Results on AlexNet and ResNet}

\begin{figure*}[htbp]
    \centering
    \includegraphics[width=0.99\linewidth]{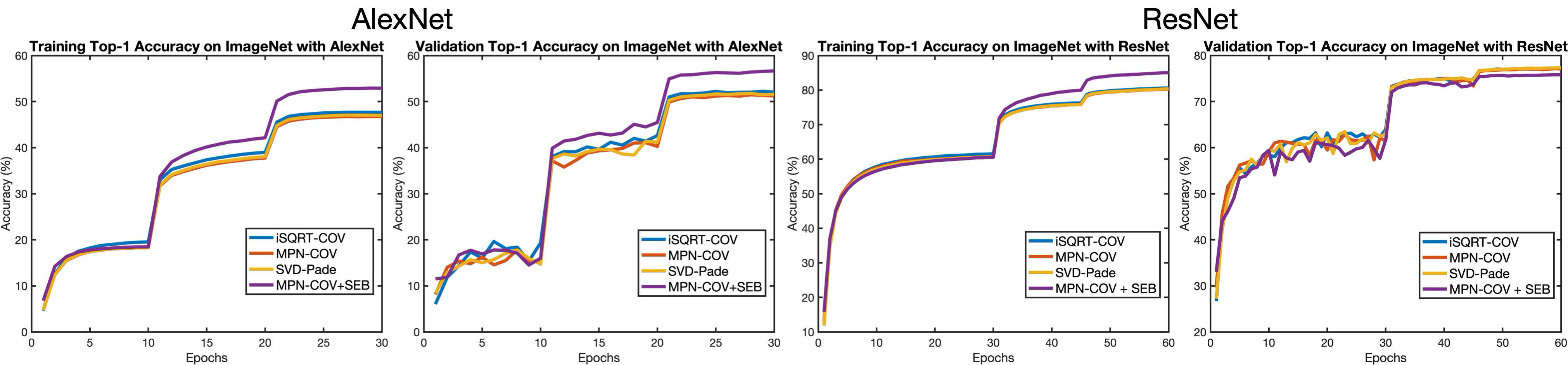}
    \caption{Training and validation top-1 accuracy of our proposed SEB and other GCP methods with AlexNet and ResNet architectures on ImageNet.}
    \label{fig:alex_res_net}
\end{figure*}

Table~\ref{tab:performances_imagenet} shows the validation top-1 and top-5 accuracy on ImageNet. For AlexNet, our proposed SEB significantly improves the performances of MPN-COV~\cite{li2017second} by about $5$ \%. However, it brings $1$ \% performance drop on ResNet-50 though the training accuracy still surpass others by a large margin (see Fig.~\ref{fig:alex_res_net}). This phenomenon implies that the ResNet-50 equipped with our SEB has the risks of over-fitting and the poorer generalization performances. We observe that the eigenvalue distributions of the GCP methods on ImageNet are quite different from those on fine-grained benchmarks. For ImageNet, both the overall eigenvalue range and the eigenvalue differences are much smaller. Consider the large class complexity and the considerable amount of noise of ImageNet~\cite{li2017webvision,beyer2020we}, the small eigenvalues do not necessarily capture the class-specific features and the subtle classification clues. On the contrary, the small eigenvalues are more likely to encode only the data noise. Therefore, amplifying the insignificant eigenvalues might bring side effects on the generalization ability. To validate this guess, we also apply our proposed explainability methods on the ResNet-50 of MPN-COV~\cite{li2017second} trained on ImageNet to evaluate the behavior of eigenvalues, which will be illustrated in the following sections. Despite the analyses from the perspective of visual explainability, we think this problem is worth further research in our future work.

\begin{table}[htbp]
\caption{Comparison on validation accuracy with other GCP methods on ImageNet.} 
    \centering
    \resizebox{0.99\linewidth}{!}{
    \setlength{\tabcolsep}{1pt}
    \begin{tabular}{r|c|c|c|c}
    \toprule
        \multirow{2}*{Method} & \multicolumn{2}{c|}{AlexNet~\cite{krizhevsky2017imagenet}} & \multicolumn{2}{c}{ResNet-50~\cite{he2016deep}} \\
        \cline{2-5}
        & top-1 acc (\%) & top-5 acc (\%) & top-1 acc (\%) & top-5 acc (\%) \\
        \hline
        iSQRT-COV~\cite{li2018towards} & 52.06 & 66.36 & 77.19 & 93.40\\
        MPN-COV~\cite{li2017second}  &51.23 &75.72 &77.07 &93.25 \\
        SVD-Pad\'e~\cite{song2021approximate} &51.59 &76.09  & \textbf{77.33} & \textbf{93.49} \\
        MPN-COV + Our SEB  & \textbf{56.68} ($\uparrow$ 5.45) & \textbf{79.94} ($\uparrow$ 4.22) & 75.79 ($\downarrow$ 1.27 ) & 92.58 ($\downarrow$ 0.67)\\
    \bottomrule
    \end{tabular}
    }
    \label{tab:performances_imagenet}
\end{table}

\subsection{Backward-based Explainability on ImageNet}

\begin{figure*}
    \centering
    \includegraphics[width=0.99\linewidth]{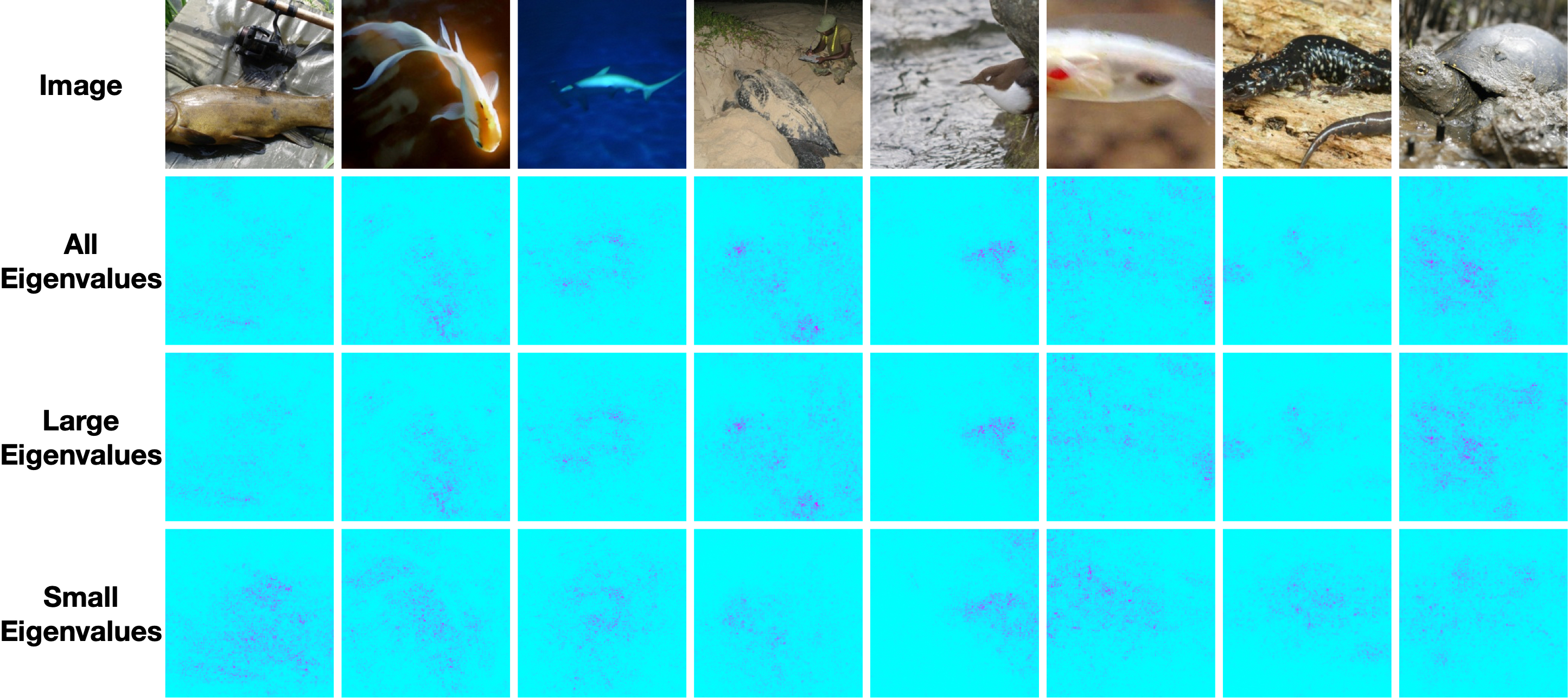}
    \caption{Backward-based visual explanations of different eigenvalues on ImageNet. Different from the situation of fine-grained benchmarks, the large eigenvalues have almost identical input activations with all the eigenvalues. By contrast, the small eigenvalues exhibit less similar input responses.}
    \label{fig:imagenet_backward}
\end{figure*}

\begin{table}
\caption{The average correlation coefficient and the MAE between the input responses of the specific eigenvalues and the responses of all the eigenvalues. Here we use vanilla ReLU back-propagation rule~\cite{simonyan2013deep}. The evaluation is conducted on the subset of randomly selected $2,000$ images from the ImageNet validation set.}
\label{tab:imagenet_similarity}
\centering
\resizebox{0.99\linewidth}{!}{
\begin{tabular}{c|c|c}\toprule  
Eigenvalue & {Correlation Coefficient} & {Mean Absolute Error}\\ 
\hline
Large &\textbf{1.00} & \textbf{2.1e-8}\\  
Small &0.68 & 2.8e-2\\
\bottomrule
\end{tabular}
}
\end{table} 
Fig.~\ref{fig:imagenet_backward} shows some examples of input responses to different eigenvalues. Unlike the case of fine-grained benchmarks, the large eigenvalues almost generate the identical input activations with all the eigenvalues, while the visualizations of small eigenvalues are less similar. Table~\ref{tab:imagenet_similarity} compares the quantitative evaluation results. For the large eigenvalues, the MAE is nearly negligible and the correlation coefficient achieves the maximum. This demonstrate that on ImageNet the decisions of the GCP networks are almost entirely dependent on the large eigenvalues and the associated eigenvectors.

\subsection{Perturbation-based Explainability on ImageNet}

\begin{figure*}
    \centering
    \includegraphics[width=0.99\linewidth]{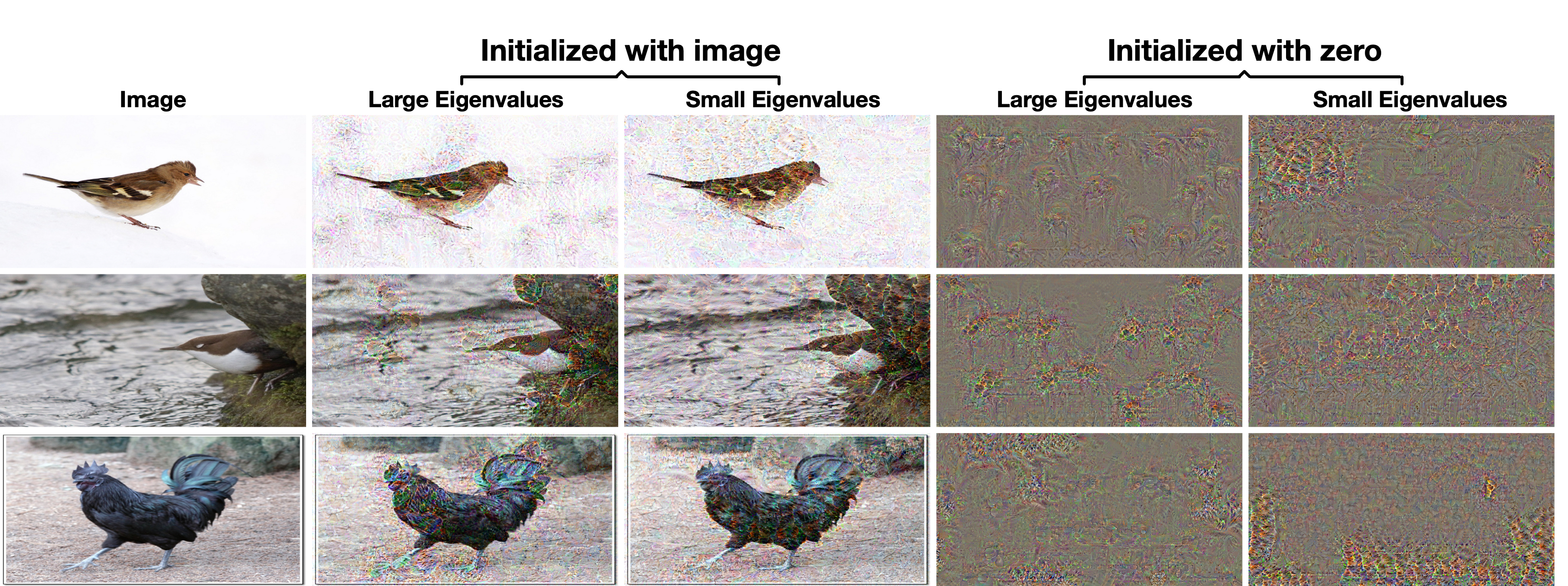}
    \caption{Perturbation-based visualization on ImageNet. Unlike the case of fine-grained benchmarks, the small eigenvalues do not show obviously class-relevant feature patterns.}
    \label{fig:imagenet_perturb}
\end{figure*}

Fig.~\ref{fig:imagenet_perturb} shows several examples of perturbation-based visualizations on ImageNet. Different from the situation on fine-grained benchmarks, the learnt feature patterns of the small eigenvalues are not obviously class-relevant. In contrast, the large eigenvalues show some signs of the class-specific feature patterns.

\subsection{FGVC Performance with GCP ResNet-50}

\begin{table}[h]
\vspace{-0.1cm}
\caption{Performances on fine-grained benchmarks when the backbone is GCP ResNet-50. GCP ResNet-50 means that the ResNet-50 equipped with the GCP meta-layer is first trained on ImageNet from scratch. GAP refers to the Global Average Pooling layer used in the ResNet architecture. For MPN-COV~\cite{li2017second} and iSQRT-COV~\cite{li2018towards}, the results are slightly different from the values reported in the original papers as different platforms and training protocols are used.}
    \centering
    \resizebox{0.99\linewidth}{!}{
    \label{tab:gcp_performances}
    \setlength{\tabcolsep}{1pt}
    \begin{tabular}{r|c|c|c|c|c}
    \toprule
        \multicolumn{2}{c|}{Backbone}  & \multirow{2}*{Method} & \multirow{2}*{Birds~\cite{WelinderEtal2010}} & \multirow{2}*{Aircrafts~\cite{maji2013fine}} & \multirow{2}*{Cars~\cite{KrauseStarkDengFei-Fei_3DRR2013}}  \\
        \cline{1-2}
        Architecture & Pooling Layer& & & &\\
        \hline
        \multirow{2}*{GCP ResNet-50} & \multirow{2}*{$2^{nd}$-order GCP}& iSQRT-COV~\cite{li2018towards} &\textbf{87.3} &89.5 & 91.7 \\
        &&MPN-COV~\cite{li2017second} &87.2 &90.5 &92.8 \\
        \hline
        \multirow{3}*{ResNet-50~\cite{he2016deep}} & \multirow{3}*{$1^{st}$-order GAP} &iSQRT-COV~\cite{li2018towards} &84.7 &89.6 &91.4 \\
       & &MPN-COV~\cite{li2017second} &84.3 &89.9 &91.7 \\
       & &MPN-COV {+} Our SEB & {86.2} ($\downarrow$ 1.0) &\textbf{91.4} ($\uparrow$ 0.9) & \textbf{93.6}  ($\uparrow$ 0.8)\\
    \bottomrule
    \end{tabular}
    }
\end{table}

Both MPN-COV~\cite{li2017second} and iSQRT-COV~\cite{li2018towards} first train ResNet-50 models with each GCP meta-layer on ImageNet from scratch and then use the same model to fine-tune on the fine-grained benchmarks. Although their reported results on fine-grained benchmarks are claimed to be based on the backbone ResNet-50, strictly speaking their backbone is actually ResNet-50 with the GCP meta-layer rather than the ordinary widely used ResNet-50. Table~\ref{tab:gcp_performances} shows their performances on these two backbones. As can be observed, only when the backbone equipped with the GCP meta-layer is trained on ImageNet from scratch, the GCP methods MPN-COV~\cite{li2017second} and iSQRT-COV~\cite{li2018towards} can achieve comparable performances on fine-grained benchmarks. This may limit their practical usage as a dedicated backbone with the GCP layer is needed. Our proposed SEB is free of this constraint. It does not pose any backbone requirement and can achieve impressive performances on the commonly used backbones.


\end{document}